%% file: main.tex
\documentclass[11pt]{article}

\usepackage[preprint]{acl}

\usepackage{times}
\usepackage{latexsym}
\usepackage{algorithm}
\usepackage{algpseudocode}
\usepackage{xcolor}
\usepackage{amsmath}
\usepackage{booktabs} 
\usepackage{amsmath}
\usepackage{amssymb} 
\usepackage{float}
\usepackage[most]{tcolorbox}
\usepackage{xurl}
\usepackage{hyperref}
\usepackage{multirow}
\usepackage{colortbl} 
\usepackage{pgfplots}\pgfplotsset{compat=1.17}
\usepackage{subcaption}
\usepackage{stfloats}
\usepackage{fix-cm}
\usepackage{silence}
\usetikzlibrary{arrows.meta}
\newcommand{\refthm}[2]{\hyperref[#1]{Theorem~#2}}

\newcommand{\lcomment}[1]{%
  \State \textcolor{gray!80}{$\triangleright$ {%
    \itshape
    \renewcommand{\texttt}[1]{{\normalfont\ttfamily ##1}}%
    #1%
  }}%
}

\usepackage[T1]{fontenc}
\DeclareFontShape{T1}{ptm}{m}{scit}{<->ssub*ptm/m/sc}{}
\DeclareFontShape{T1}{ptm}{bx}{scit}{<->ssub*ptm/bx/sc}{}

\usepackage[utf8]{inputenc}

\usepackage{microtype}

\usepackage{inconsolata}

\usepackage{graphicx}

\title{Streaming Communication in Multi-Agent Reasoning}

\author{
 \textbf{Zhen Yang\textsuperscript{1}},
 \textbf{Xiaogang Xu\textsuperscript{3}},
 \textbf{Wen Wang\textsuperscript{3}},
 \textbf{Cong Chen\textsuperscript{3}},
\\
 \textbf{Xander Xu\textsuperscript{2}\thanks{Co-corresponding authors.}},
 \textbf{Ying-Cong Chen\textsuperscript{1,4}\footnotemark[\value{footnote}]}
\\
 \textsuperscript{1}HKUST(GZ),
 \textsuperscript{2}Alibaba Group,
 \textsuperscript{3}ZJU,
 \textsuperscript{4}HKUST
\\
 \small{
  \href{zheny.cs@gmail.com}{zheny.cs@gmail.com}, \href{yingcongchen@ust.hk}{yingcongchen@ust.hk}
 }
}

\begin{document}
\maketitle
\input{latex/secs/01_abstract}
\input{latex/secs/02_introduction}
\input{latex/secs/03_related_work}

\input{latex/secs/04_method}

\input{latex/secs/05_results}

\input{latex/secs/06_conclusion}
\input{latex/secs/07_limitation}

\bibliography{custom}
\appendix
\input{latex/secs/08_appendix}

\end{document}

%% file: latex/secs/01_abstract.tex
\begin{abstract}

Multi-agent reasoning systems adopt a ``\emph{generate-then-transfer}''
paradigm that forces end-to-end latency to scale linearly with pipeline depth.
We introduce \textsc{StreamMA}, a multi-agent reasoning system that streams each \emph{reasoning step} to downstream agents as soon as it is generated, pipelining adjacent agents and thus reducing latency.
Surprisingly, this pipelining also improves effectiveness: because multi-step reasoning quality is non-uniform and early steps are more reliable than later ones, working with these reliable early steps instead of the full chain prevents error-prone late steps from misleading downstream agents.
We formalize both advantages with the first closed-form joint analysis of stream, serial, and single protocols, deriving the effectiveness ordering, speedup upper bound, and cost ratio.
Across eight reasoning benchmarks spanning mathematics, science, and code, three frontier LLMs (Claude Opus~4.6, GPT-5.4, and GLM-5.2), and three topologies (Chain, Tree, Graph),
\textsc{StreamMA} outperforms both baselines
(avg.\ \textbf{+7.3\,pp}, max \textbf{+22.4\,pp} on HMMT\,2026; Claude Opus~4.6-High).
Beyond these contributions, we discover a ``\emph{step-level scaling law}'': increasing per-agent steps
consistently improves both effectiveness and efficiency, a new scaling dimension orthogonal to and composable with agent-count scaling.
Our \href{https://zhenyangcs.github.io/StreamMA-website/}{\textcolor{blue}{project page}} and \href{https://github.com/EnVision-Research/StreamMA}{\textcolor{blue}{code}} are available.
\end{abstract}

%% file: latex/secs/02_introduction.tex
\section{Introduction}

\begin{figure}[t]
  \centering
  \includegraphics[width=1.0\linewidth]{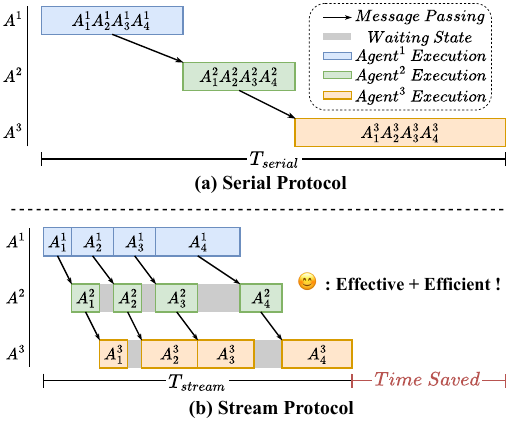}
\caption{\textbf{Communication protocols.} \textbf{(a)~Serial}: the downstream agent receives the upstream agent's complete response before execution. \textbf{(b)~Stream}: the downstream agent receives each upstream reasoning step as it is generated, enabling pipelined execution.}  \label{fig:method}
\end{figure}

Multi-agent systems have emerged as a paradigm for complex reasoning: agents are organized as nodes of a directed acyclic graph (DAG), with each directed edge governing information propagation from one agent to its successor, leveraging specialization and cross-verification to surpass a single model~\citep{autogen,metagpt,chatdev}, with growing benefits as the number of agents increases~\citep{scaling}.
Yet existing frameworks share the same communication assumption: an upstream agent must finish its entire response before passing it downstream.
This ``generate-then-transfer'' (Serial, Fig.~\ref{fig:method}(a)) protocol forces downstream agents to remain idle and latency to scale with pipeline depth.
While streaming partial outputs is standard within a single LLM~\citep{static}, between agents the transfer unit has remained a complete response, and its effect on effectiveness remains underexplored.

A natural question arises: can we forward partial outputs before the upstream agent finishes?
Conventional wisdom holds that incomplete context must hurt downstream quality: ``more information yields better decisions.''
Yet this intuition does not always hold: we find that in multi-step reasoning, receiving less context can improve downstream reasoning quality.
We propose \textsc{StreamMA} (Fig.~\ref{fig:method}(b)), a multi-agent reasoning system built on \emph{Stream}, a reasoning-step-level communication protocol that shifts the transmission unit from complete responses to reasoning steps, forwarding each step once produced, pipelining upstream and downstream agents.
Across eight benchmarks in mathematics, science, and code, \textsc{StreamMA} outperforms Serial by an average $+7.3$\,pp (peak $+22.4$\,pp on HMMT\,2026, Claude Opus~4.6-High) while reducing latency: less waiting, yet better reasoning.

Why does the downstream agent benefit from only the early reasoning steps more than the full response?
On complex reasoning tasks, step-level quality is position-dependent: early steps tend to be reliable, whereas later steps degrade.
The Serial protocol forces the downstream agent to condition on the entire response, including its error-prone late steps.
Stream lets it begin from the most reliable prefix; by the time degraded steps arrive, the agent has formed its own independent reasoning trajectory, diluting their impact.
Controlled step-level perturbations cross-validate this asymmetry: corrupting only the tail steps of an upstream reasoning trajectory leaves Stream nearly intact and ahead of Serial by $+24.0$\,pp, while corrupting only its head steps flips the gap to $-36.0$\,pp (Sec.~\ref{sec:stepwise_perturb}).
We formalize this mechanism: \refthm{thm:1}{1} gives an effectiveness ordering across all six regimes, predicting Stream optimal when early steps are reliable and late ones degrade; \refthm{thm:2}{2} and \refthm{thm:3}{3} derive the speedup upper bound and exact cost ratio under LLM serving conditions.

Beyond mechanism and theory, our experiments uncover a second scaling axis, orthogonal to the well-studied agent-count axis: at fixed agent count $A$, increasing per-agent steps $S$ consistently improves effectiveness and speedup, defining a step-level scaling law fully compositional with agent-count scaling. On HMMT\,2026 with GPT-5.4-None, scaling $A$ alone from $2$ to $64$ lifts effectiveness from $58.3\%$ to $68.2\%$; further increasing $S$ to $64$ pushes it to $73.5\%$ with $26.9\times$ wall-clock speedup ($83\%$ of the \refthm{thm:2}{2} upper bound).
\textsc{StreamMA} thus simultaneously serves as a reasoning enhancer and an inference accelerator.

In summary, our contributions are as follows:
\begin{itemize}
\item \textbf{Protocol.}
  We introduce Stream, a reasoning-step-level communication protocol that forwards each reasoning step upon completion, pipelining upstream and downstream agents over arbitrary DAG topologies. Stream improves effectiveness while reducing latency.
\item \textbf{Theory.}
  We present the first closed-form joint analysis of Stream, Serial, and Single protocols: \refthm{thm:1}{1} provides an effectiveness ordering, \refthm{thm:2}{2} a speedup upper bound, and \refthm{thm:3}{3} an exact cost ratio.
\item \textbf{Empirics.}
  Across eight benchmarks, three LLMs, and three topologies, \textsc{StreamMA} consistently outperforms both baselines in average effectiveness (Claude $+7.3$\,pp over Serial, $+16.3$\,pp over Single; GPT $+1.5$\,pp over Serial, $+4.9$\,pp over Single; GLM $+7.2$\,pp over Serial, $+20.0$\,pp over Single).
\item \textbf{Discovery.}
  We discover a step-level scaling law under Stream: at fixed agent count $A$, increasing per-agent steps~$S$ consistently improves effectiveness and speedup. At $A{=}64$, scaling $S$ to $64$ gains $+5.3$\,pp complementary to agent-count scaling, with $26.9\times$ speedup.
\end{itemize}

%% file: latex/secs/03_related_work.tex
\section{Related Work}

\paragraph{Multi-Agent Reasoning and Communication.}
Decomposing reasoning across collaborating agents has become a mainstream paradigm for complex LLM tasks~\citep{autogen,metagpt,chatdev,camel,improving,chen2024agentverse,wang2025mixture}, with prior work advancing it along three orthogonal axes: the communication topology~\citep{scaling,zhuge2024gptswarm,liu2024dynamic,zhang2025aflow}; the content exchanged between agents, such as intermediate rationales~\citep{yin2023exchange} or KV-cache representations~\citep{fu2025cache}, and how much of it~\citep{rizvi2025benefits}; and the scale of agents~\citep{scaling,li2024more}.
Yet all three share a generate-then-transfer assumption: an upstream agent must finalize its response before any downstream agent can act, forcing per-round sequential waits and forgoing pipeline parallelism.
We instead refine the granularity of communication, from full responses to reasoning steps. This shift exposes a new design axis: per-agent step count, orthogonal to agent-count scaling. Increasing per-agent steps simultaneously improves effectiveness and speedup, termed the step-level scaling law.

\paragraph{Step-Level Reasoning Quality.}
Step-by-step reasoning is now standard~\citep{cot,tot}, with iterative-refinement variants such as Self-Refine~\citep{selfrefine}, Reflexion~\citep{reflexion}, and multi-agent debate~\citep{improving} pushing output quality higher.
On complex reasoning tasks, step quality is position-dependent: \citet{wu2025more} show that CoT accuracy peaks at an optimal length and degrades beyond it, and process supervision~\citep{uesato2022solving,lightman2024let,wang2024math} scores reliability step by step. \citet{MTI} find that uncertainty is non-uniform even at the token level. So far, this property has been exploited only within a single model's reasoning chain, never to design the inter-agent protocol.
We are the first to lift this property into protocol design: our \refthm{thm:1}{1} gives a closed-form effectiveness ordering of Single, Serial, and Stream, with explicit conditions under which each mode is provably optimal.

\begin{algorithm}[t]
\caption{\textsc{Serial Execution}}
\label{alg:serial}
\begin{algorithmic}[1]
\Require $Q$; $(\mathrm{Agent}^a, \mathrm{ctx}_a, \mathrm{queue}_a)_{a=1}^{A}$
\Statex $\circ$ \textit{$Q$: query}
\Statex $\circ$ \textit{$\mathrm{ctx}_a$: per-agent context}
\Statex $\circ$ \textit{$\mathrm{queue}_a$: FIFO queue}
\Statex $\circ$ \textit{chain: $\mathrm{Agent}^1 \to \cdots \to \mathrm{Agent}^A$}
\State $\mathit{msg} \gets Q$
\For{$a = 1$ \textbf{to} $A$}
    \State $\mathrm{ctx}_a.\mathrm{append}(\mathit{msg})$
    \lcomment{wait; complete output}
    \State $\mathit{msg} \gets \mathrm{LLM}(\mathrm{ctx}_a)$
\EndFor
\end{algorithmic}
\end{algorithm}

\paragraph{Pipeline Parallelism and Streaming Inference.}
Pipeline parallelism is classical in distributed training~\citep{narayanan2021efficient}, and streaming inference for LLMs~\citep{static} has been pursued along two axes: at the intra-agent level, via speculative decoding~\citep{speculativedecoding,accelerating,cai2024medusa}, Group Think~\citep{hsu2025group}, and Multi-Stream LLMs~\citep{su2026multi}; and at the inter-agent level, via skeleton expansion~\citep{ning2024skeleton}, speculative agent actions~\citep{ye2025speculative}, and staircase streaming~\citep{wang2025staircase}.
Yet across all of these, streaming primarily serves as a speedup mechanism; any effectiveness gains are side-effects rather than the central design principle.
Our Stream instead operates at the reasoning-step level over arbitrary multi-agent DAGs: a granularity coarser than tokens or fixed token chunks, yet finer than pre-decomposed skeletons or speculative agent actions, and unconstrained by topology. To our knowledge, we are the first to show that it can improve effectiveness, not just accelerate it (\refthm{thm:1}{1}). At the inter-agent level, our protocol is orthogonal to all the intra-agent methods discussed above.

%% file: latex/secs/04_method.tex
\begin{algorithm}[t]
\caption{\textsc{Stream Execution}}
\label{alg:stream}
\begin{algorithmic}[1]
\State $\mathrm{queue}_1.\mathrm{put}(Q)$
\lcomment{all agents concurrent}
\For{$a = 1$ \textbf{to} $A$ \textbf{in parallel}}
    \While{$\mathit{msg} \gets \mathrm{queue}_a.\mathrm{get}()$}
        \State $\mathrm{ctx}_a.\mathrm{append}(\mathit{msg})$
        \lcomment{yield step-by-step}
        \State $\mathit{steps} \gets \mathrm{LLM}(\mathrm{ctx}_a,\,\texttt{stream=True})$
        \For{\textbf{each} $\mathit{step}$ \textbf{from} $\mathit{steps}$}
            \If{$a < A$}
                \lcomment{push; no wait}
                \State $\mathrm{queue}_{a+1}.\mathrm{put}(\mathit{step})$
            \EndIf
            \lcomment{KV cache reuse}
            \State $\mathrm{ctx}_a.\mathrm{append}(\mathit{step})$
        \EndFor
    \EndWhile
\EndFor
\end{algorithmic}
\end{algorithm}

\section{Method}

Since chains admit closed-form analysis and naturally compose general DAGs, we develop our theory on a chain of $A$ agents $(\mathrm{Agent}^1 \to \cdots \to \mathrm{Agent}^A)$ each producing $S$ reasoning steps. The results readily extend to arbitrary DAGs. We first describe the algorithm (Sec.~\ref{sec:Algorithm Description}), then characterize its effectiveness (Sec.~\ref{sec:effectiveness}) and efficiency (Sec.~\ref{sec:efficiency}). Notation is summarized in App.~\ref{sec:notation}. We compare three execution modes throughout: \textbf{Stream}, our stream protocol; \textbf{Serial}, the generate-then-transfer serial protocol; and \textbf{Single}, the single-agent protocol. We refer to our framework, including the protocol and its theoretical analysis, as \textsc{StreamMA}.

\subsection{Algorithm Description}
\label{sec:Algorithm Description}

Alg.~\ref{alg:serial} and Alg.~\ref{alg:stream} contrast the serial and stream protocols. The key difference lies in the timing of message passing: in the serial protocol, each agent waits for the blocking call $\mathrm{LLM}(\mathrm{ctx}_a)$ to return the complete response (Alg.~\ref{alg:serial}, line~5) before the next agent can begin; in the stream protocol, each agent issues a streaming call that yields steps one by one (Alg.~\ref{alg:stream}, lines~7--8) and immediately pushes every completed step to $\mathrm{queue}_{a+1}$ (line~11) without waiting for subsequent steps. Step boundaries are explicit delimiters that the prompts in App.~\ref{sec:prompts} instruct the model to emit. All agents execute concurrently (line~3), enabling $\mathrm{Agent}^{a+1}$ to process step $s$ while $\mathrm{Agent}^a$ is still generating step $s+1$, forming pipeline parallelism. Moreover, each downstream agent ($a \geq 2$) is called $S$ times, with prior steps serving as context for each subsequent call (line~14); the appended context naturally forms a shared prefix amenable to KV-cache reuse.

The chain topology generalizes to arbitrary DAGs by (i)~broadcasting $Q$ to all source nodes (in-degree~0) and (ii)~pushing each step to all direct successors; multi-predecessor nodes process incoming steps immediately upon arrival, without any synchronization barrier across predecessors. The full pseudocode is given in App.~\ref{sec:stream-dag-extension}.

\subsection{Effectiveness Characterization}
\label{sec:effectiveness}

\textsc{StreamMA} consistently outperforms the Serial and Single baselines in our experiments. We now characterize when and why: we model each upstream step $j$ ($1 \le j \le S$) as correct with probability $p_j$; a correct step raises downstream step-level correctness by $\delta>0$ and an incorrect step lowers it by $\varepsilon>0$. \textit{For clarity we assume uniform $\delta,\varepsilon$ across positions; the position-dependent $\delta_j,\varepsilon_j$ case yields the same results in vector form.} The expected step-level correctness change from step $j$ is $\mu_j = p_j\delta - (1-p_j)\varepsilon$, which is non-negative iff $p_j \geq \varepsilon/(\delta+\varepsilon) \triangleq p^*$. We define $\mathrm{sCorr}^{\mathrm{mode}}$ ($\mathrm{mode}\!\in\!\{\mathrm{stream},\mathrm{serial},\mathrm{single}\}$) as the mean of $\{\mu_j\}$ under each mode; \textit{sCorr is positively correlated with task-level accuracy}. Three weighted means of $\{p_j\}$, the uniform mean $\bar{p}$, the head-weighted $p_{\mathrm{head}}$ (earlier steps weighted more), and the tail-weighted $p_{\mathrm{tail}}$ (later steps weighted more), govern the sCorr ordering among the three modes.

\begin{figure*}[t]
\begin{tcolorbox}[colback=green!4!white, colframe=green!45!black, arc=3pt, boxrule=0.6pt, fontupper={\fontsize{9.5pt}{11.4pt}\linespread{1.3}\selectfont}]
\label{thm:1}\textbf{Theorem 1} (Effectiveness Ordering). \textit{Depending on how $\bar{p}$, $p_{\mathrm{head}}$, $p_{\mathrm{tail}}$ (uniform, head-weighted, and tail-weighted step-correctness means) compare to $p^*$, the sCorr ordering among three modes falls into six cases:}

\noindent\tikz\draw[gray,dashed](0,0)--(\linewidth,0);

\noindent\textbf{(I) Stream advantage} [$p_{\mathrm{head}}\!>\!p^*$ and $p_{\mathrm{tail}}\!<\!p^*$]\textbf{:}

\noindent
\begin{minipage}[t]{0.48\linewidth}
\noindent\begin{minipage}[c]{0.72\linewidth}
\textit{(a)~If $\bar{p} > p^*$:}\par
\scalebox{0.88}{$\mathrm{sCorr}^{\mathrm{stream}} > \mathrm{sCorr}^{\mathrm{serial}} > \mathrm{sCorr}^{\mathrm{single}}$}
\end{minipage}%
\hfill
\begin{minipage}[c]{0.25\linewidth}
\centering
\begin{tikzpicture}[scale=0.32,>=stealth]
\draw[->] (0,0) -- (3.3,0) node[right,font=\tiny] {$j$};
\draw[->] (0,0) -- (0,2.2) node[above,font=\tiny] {$p_j$};
\draw[dashed,gray!60] (0,0.9) -- (3.1,0.9) node[right,font=\tiny] {$p^*$};
\draw[blue!75!black,thick] (0.4,1.7)--(1.0,1.7)--(1.6,1.7)--(2.2,0.3)--(2.8,0.3);
\foreach \x in {0.4,1.0,1.6} \filldraw[green!55!black] (\x,1.7) circle (2pt);
\foreach \x in {2.2,2.8} \filldraw[red!65!black] (\x,0.3) circle (2pt);
\end{tikzpicture}
\end{minipage}
\end{minipage}%
\hfill
\begin{minipage}[t]{0.48\linewidth}
\noindent\begin{minipage}[c]{0.72\linewidth}
\textit{(b)~If $\bar{p} < p^*$:}\par
\scalebox{0.88}{$\mathrm{sCorr}^{\mathrm{stream}} > \mathrm{sCorr}^{\mathrm{single}} > \mathrm{sCorr}^{\mathrm{serial}}$}
\end{minipage}%
\hfill
\begin{minipage}[c]{0.25\linewidth}
\centering
\begin{tikzpicture}[scale=0.32,>=stealth]
\draw[->] (0,0) -- (3.3,0) node[right,font=\tiny] {$j$};
\draw[->] (0,0) -- (0,2.2) node[above,font=\tiny] {$p_j$};
\draw[dashed,gray!60] (0,0.9) -- (3.1,0.9) node[right,font=\tiny] {$p^*$};
\draw[blue!75!black,thick] (0.4,1.7)--(1.0,1.7)--(1.6,0.3)--(2.2,0.3)--(2.8,0.3);
\foreach \x in {0.4,1.0} \filldraw[green!55!black] (\x,1.7) circle (2pt);
\foreach \x in {1.6,2.2,2.8} \filldraw[red!65!black] (\x,0.3) circle (2pt);
\end{tikzpicture}
\end{minipage}
\end{minipage}

\noindent\tikz\draw[gray,dashed](0,0)--(\linewidth,0);

\noindent
\begin{minipage}[t]{0.48\linewidth}
\textbf{(II) Serial advantage} [$\bar{p}\!>\!p^*$ and $p_{\mathrm{tail}}\!>\!p^*$]\textbf{:}

\noindent\begin{minipage}[c]{0.72\linewidth}
\textit{(a)~If $p_{\mathrm{head}} > p^*$:}\par
\scalebox{0.88}{$\mathrm{sCorr}^{\mathrm{serial}} > \mathrm{sCorr}^{\mathrm{stream}} > \mathrm{sCorr}^{\mathrm{single}}$}
\end{minipage}%
\hfill
\begin{minipage}[c]{0.25\linewidth}
\centering
\begin{tikzpicture}[scale=0.32,>=stealth]
\draw[->] (0,0) -- (3.3,0) node[right,font=\tiny] {$j$};
\draw[->] (0,0) -- (0,2.2) node[above,font=\tiny] {$p_j$};
\draw[dashed,gray!60] (0,0.9) -- (3.1,0.9) node[right,font=\tiny] {$p^*$};
\draw[blue!75!black,thick] (0.4,1.7)--(1.0,1.7)--(1.6,1.7)--(2.2,1.7)--(2.8,1.7);
\foreach \x in {0.4,1.0,1.6,2.2,2.8} \filldraw[green!55!black] (\x,1.7) circle (2pt);
\end{tikzpicture}
\end{minipage}

\noindent\begin{minipage}[c]{0.72\linewidth}
\textit{(b)~If $p_{\mathrm{head}} < p^*$:}\par
\scalebox{0.88}{$\mathrm{sCorr}^{\mathrm{serial}} > \mathrm{sCorr}^{\mathrm{single}} > \mathrm{sCorr}^{\mathrm{stream}}$}
\end{minipage}%
\hfill
\begin{minipage}[c]{0.25\linewidth}
\centering
\begin{tikzpicture}[scale=0.32,>=stealth]
\draw[->] (0,0) -- (3.3,0) node[right,font=\tiny] {$j$};
\draw[->] (0,0) -- (0,2.2) node[above,font=\tiny] {$p_j$};
\draw[dashed,gray!60] (0,0.9) -- (3.1,0.9) node[right,font=\tiny] {$p^*$};
\draw[blue!75!black,thick] (0.4,0.3)--(1.0,0.3)--(1.6,1.7)--(2.2,1.7)--(2.8,1.7);
\foreach \x in {0.4,1.0} \filldraw[red!65!black] (\x,0.3) circle (2pt);
\foreach \x in {1.6,2.2,2.8} \filldraw[green!55!black] (\x,1.7) circle (2pt);
\end{tikzpicture}
\end{minipage}
\end{minipage}%
\hfill
\begin{minipage}[t]{0.48\linewidth}
\textbf{(III) Single advantage} [$p_{\mathrm{head}}\!<\!p^*$ and $\bar{p}\!<\!p^*$]\textbf{:}

\noindent\begin{minipage}[c]{0.72\linewidth}
\textit{(a)~If $p_{\mathrm{tail}} < p^*$:}\par
\scalebox{0.88}{$\mathrm{sCorr}^{\mathrm{single}} > \mathrm{sCorr}^{\mathrm{stream}} > \mathrm{sCorr}^{\mathrm{serial}}$}
\end{minipage}%
\hfill
\begin{minipage}[c]{0.25\linewidth}
\centering
\begin{tikzpicture}[scale=0.32,>=stealth]
\draw[->] (0,0) -- (3.3,0) node[right,font=\tiny] {$j$};
\draw[->] (0,0) -- (0,2.2) node[above,font=\tiny] {$p_j$};
\draw[dashed,gray!60] (0,0.9) -- (3.1,0.9) node[right,font=\tiny] {$p^*$};
\draw[blue!75!black,thick] (0.4,0.3)--(1.0,0.3)--(1.6,0.3)--(2.2,0.3)--(2.8,0.3);
\foreach \x in {0.4,1.0,1.6,2.2,2.8} \filldraw[red!65!black] (\x,0.3) circle (2pt);
\end{tikzpicture}
\end{minipage}

\noindent\begin{minipage}[c]{0.72\linewidth}
\textit{(b)~If $p_{\mathrm{tail}} > p^*$:}\par
\scalebox{0.88}{$\mathrm{sCorr}^{\mathrm{single}} > \mathrm{sCorr}^{\mathrm{serial}} > \mathrm{sCorr}^{\mathrm{stream}}$}
\end{minipage}%
\hfill
\begin{minipage}[c]{0.25\linewidth}
\centering
\begin{tikzpicture}[scale=0.32,>=stealth]
\draw[->] (0,0) -- (3.3,0) node[right,font=\tiny] {$j$};
\draw[->] (0,0) -- (0,2.2) node[above,font=\tiny] {$p_j$};
\draw[dashed,gray!60] (0,0.9) -- (3.1,0.9) node[right,font=\tiny] {$p^*$};
\draw[blue!75!black,thick] (0.4,0.3)--(1.0,0.3)--(1.6,0.3)--(2.2,1.7)--(2.8,1.7);
\foreach \x in {0.4,1.0,1.6} \filldraw[red!65!black] (\x,0.3) circle (2pt);
\foreach \x in {2.2,2.8} \filldraw[green!55!black] (\x,1.7) circle (2pt);
\end{tikzpicture}
\end{minipage}
\end{minipage}
\end{tcolorbox}
\end{figure*}

The six examples in \refthm{thm:1}{1} fix $p^*$ only for simplicity. ``Strong'' and ``weak'' describe $p_j$ relative to this threshold, which may vary by step.

\noindent\textbf{Stream advantage.} Cases~I.a and~I.b share the error-accumulation pattern in LLM reasoning: early steps are reliable but late ones fall below $p^*$. When context arrives matters more than how much: under Stream, the downstream agent receives early steps and begins reasoning; by the time late steps arrive, their impact is diluted. Under Serial, it must read the entire response, mixing clean early reasoning with error-prone late steps. In Case~I.a, $\bar{p} > p^*$ means the response is on average helpful, so Stream leads while Serial beats Single. In Case~I.b, only early steps exceed $p^*$, so Stream exposes these steps and beats Single, but Serial accumulates below-$p^*$ late steps and loses to Single.

\noindent\textbf{Serial advantage.} In Case~II.a, all steps are above $p^*$, so more context is beneficial: ``more information, better decisions'' holds. Serial sees the full response and wins; Stream sees partial but helpful steps and beats Single. In Case~II.b, early steps are below $p^*$ but late ones are accurate enough that $\bar{p} > p^*$; Serial profits and beats Single, yet Stream exposes the harmful early steps and ends up last. This captures self-correction: the upstream agent makes early mistakes but corrects them eventually.

\noindent\textbf{Single advantage.} Case~III.a corresponds to problems so hard that all steps fall below $p^*$, making every upstream step harmful; Single avoids any upstream influence and achieves the highest step correctness. In Case~III.b, the profile rises but $\bar{p} < p^*$ holds; this regime is rare, as a heavily corrupted prefix leaves too few remaining steps to lift $p_{\mathrm{tail}}$ above $p^*$. Serial loses to Single, and Stream, which receives the below-$p^*$ early steps, comes last.

Variable definitions, full proofs, and the extension to arbitrary DAG topologies are in App.~\ref{sec:detailed-effectiveness-analysis}.

\subsection{Efficiency Characterization}
\label{sec:efficiency}

\begin{tcolorbox}[colback=green!4!white, colframe=green!45!black, arc=3pt, boxrule=0.6pt, fontupper={\fontsize{9.5pt}{11.4pt}\linespread{1.3}\selectfont}]
\label{thm:2}\textbf{Theorem 2} (Speedup). \textit{The latency speedup of Stream over Serial is upper-bounded by:}
\[
\mathrm{Speedup} = \frac{A\bigl[(S+r_{po})\,r_{v_{dp}} + S\bigr]}{(S+A-1)(1 + \alpha\,r_{v_{dp}} + \beta\,r_{v_{dc}})}
\]
{\small\textit{where} $r_{v_{dp}}$: decode-to-prefill speed ratio; $r_{v_{dc}}$: decode-to-cache-read speed ratio; $r_{po}$: system-prompt-to-per-step-output length ratio; $\alpha$, $\beta$: average non-cached / cached context tokens per output token.}
\end{tcolorbox}

\paragraph{Speedup.}
In the streaming multi-agent system, the latency speedup is jointly governed by prefill, decode, and KV-cache read speeds (\refthm{thm:2}{2}). When $v_c \gg v_p \gg v_d$ (cache-read $\gg$ prefill $\gg$ decode speed), the upper bound reduces to $AS/(S+A-1)$, the maximum achievable speedup, and becomes insensitive to the KV-cache hit rate.

\begin{tcolorbox}[colback=green!4!white, colframe=green!45!black, arc=3pt, boxrule=0.6pt, fontupper={\fontsize{9.5pt}{11.4pt}\linespread{1.3}\selectfont}]
\label{thm:3}\textbf{Theorem 3} (Cost Ratio). \textit{Under the same setup as \refthm{thm:2}{2}, the cost ratio of Stream over Serial is:}
\[
\frac{\mathrm{Cost}_{\mathrm{stream}}}{\mathrm{Cost}_{\mathrm{serial}}}
= \rho \cdot \frac{r_{c_{pd}}\,(\alpha + r_{c_{cp}}\,\beta) + 1}{r_{c_{pd}}\,(1 + r_{po}/S) + 1}
\]
{\small\textit{where} $\rho$: Stream-to-Serial output length ratio; $r_{c_{pd}}$: prefill-to-decode price ratio; $r_{c_{cp}}$: cache-to-prefill price ratio; $\alpha$, $\beta$ as in \refthm{thm:2}{2}.}
\end{tcolorbox}

\paragraph{Cost.}
Although Stream increases API calls, the overall cost remains dominated by decode when the per-token prefill price $c_p$ is far below the per-token decode price $c_d$; KV-cache hits further reduce cost by replacing prefill with cache reads at price $c_c \ll c_p$. When pricing is decode-dominated ($r_{c_{pd}} \to 0$), the cost ratio (\refthm{thm:3}{3}) reduces to $\rho$: cost depends on how many tokens each decodes. Each downstream agent in Stream naturally forms a shared prefix amenable to KV-cache reuse, and modern serving stacks such as vLLM~\citep{kwon2023efficient} and SGLang~\citep{zheng2024sglang}, together with recent agentic extensions~\citep{droidspeak,dualpath,pan2025kvflow,ye2025kvcomm}, steadily raise the achievable hit rate, so our cost advantage strengthens as this infrastructure matures.

\paragraph{Example.}
Instantiating both bounds with Claude Opus~4.6 data ($v_d\approx 39$ t/s, $v_p\approx 6{,}000$ t/s\footnote{\url{https://artificialanalysis.ai/models/claude-opus-4-6/providers}}; $c_p=\$5$, $c_d=\$25$, $c_c=\$0.50$ per MTok\footnote{\url{https://www.anthropic.com/pricing}}) and $A=S=4$: under full KV-cache hits, the latency speedup upper bound is $2.30\times$ and the cost ratio is $0.925\rho$. At $\rho=1$, Stream saves $7.5\%$ over Serial. 

Full derivations, numerical examples, and extension to DAG topologies in App.~\ref{sec:detailed-speedup-analysis} and~\ref{sec:detailed-cost-analysis}.

%% file: latex/secs/05_results.tex
\section{Results}

\begin{table*}[t]
\centering
\small
\setlength{\tabcolsep}{4pt}
\renewcommand{\arraystretch}{1.15}
\resizebox{\textwidth}{!}{%
\begin{tabular}{ccc ccc cc ccc | c}
\toprule
\textbf{Model} & \textbf{Topo} & \textbf{Method} & \textbf{AIME25} & \textbf{AIME26} & \textbf{HMMT26} & \textbf{GPQA-D} & \textbf{HLE} & \textbf{LCB-G} & \textbf{LCB-E} & \textbf{LCB-T} & \textbf{Avg.} \\
\midrule
\multirow{7}{*}[-0.6em]{\rotatebox{90}{Claude Opus 4.6}}
 & -- & Single & 67.50 & 60.00 & 48.11 & 83.67 & 18.60 & 90.25 & 77.94 & 84.31 & 66.30 \\
\cmidrule{2-12}
 & \multirow{2}{*}{Chain} & Serial & 80.42 & 72.08 & 63.26 & 85.86 & 23.90 & 91.33 & 78.64 & 92.38 & 73.48 \\
 &  & \cellcolor{gray!12}\textsc{StreamMA} & \cellcolor{gray!12}\textbf{92.50} & \cellcolor{gray!12}\textbf{89.58} & \cellcolor{gray!12}\textbf{85.61} & \cellcolor{gray!12}\textbf{87.37} & \cellcolor{gray!12}\textbf{26.97} & \cellcolor{gray!12}\textbf{91.50} & \cellcolor{gray!12}\textbf{84.41} & \cellcolor{gray!12}\textbf{95.63} & \cellcolor{gray!12}\textbf{81.70} \\
\cmidrule{2-12}
 & \multirow{2}{*}{Tree} & Serial & 86.25 & 86.25 & 75.00 & 85.18 & 24.82 & 91.92 & 88.45 & 97.59 & 79.43 \\
 &  & \cellcolor{gray!12}\textsc{StreamMA} & \cellcolor{gray!12}\textbf{93.34} & \cellcolor{gray!12}\textbf{87.92} & \cellcolor{gray!12}\textbf{82.20} & \cellcolor{gray!12}\textbf{85.86} & \cellcolor{gray!12}\textbf{25.07} & \cellcolor{gray!12}\textbf{94.00} & \cellcolor{gray!12}\textbf{94.57} & \cellcolor{gray!12}\textbf{99.55} & \cellcolor{gray!12}\textbf{82.81} \\
\cmidrule{2-12}
 & \multirow{2}{*}{Graph} & Serial & 77.92 & 71.67 & 61.75 & 85.69 & 22.17 & 90.08 & 75.78 & 98.27 & 72.92 \\
 &  & \cellcolor{gray!12}\textsc{StreamMA} & \cellcolor{gray!12}\textbf{95.83} & \cellcolor{gray!12}\textbf{87.92} & \cellcolor{gray!12}\textbf{82.58} & \cellcolor{gray!12}\textbf{86.53} & \cellcolor{gray!12}\textbf{27.68} & \cellcolor{gray!12}\textbf{92.17} & \cellcolor{gray!12}\textbf{95.27} & \cellcolor{gray!12}\textbf{98.72} & \cellcolor{gray!12}\textbf{83.34} \\
\midrule
\multirow{7}{*}[-0.6em]{\rotatebox{90}{GPT-5.4}}
 & -- & Single & 55.83 & 71.25 & 40.53 & 77.95 & 12.08 & 91.08 & 92.48 & 96.68 & 67.24 \\
\cmidrule{2-12}
 & \multirow{2}{*}{Chain} & Serial & 60.00 & 70.42 & 54.55 & 75.08 & 14.66 & 90.08 & 97.43 & 99.02 & 70.16 \\
 &  & \cellcolor{gray!12}\textsc{StreamMA} & \cellcolor{gray!12}\textbf{61.25} & \cellcolor{gray!12}\textbf{72.50} & \cellcolor{gray!12}\textbf{59.10} & \cellcolor{gray!12}\textbf{80.30} & \cellcolor{gray!12}\textbf{14.94} & \cellcolor{gray!12}\textbf{91.17} & \cellcolor{gray!12}\textbf{99.30} & \cellcolor{gray!12}\textbf{99.47} & \cellcolor{gray!12}\textbf{72.25} \\
\cmidrule{2-12}
 & \multirow{2}{*}{Tree} & Serial & 59.17 & \textbf{75.83} & 56.07 & 76.77 & 14.83 & 88.33 & 93.81 & \textbf{99.25} & 70.51 \\
 &  & \cellcolor{gray!12}\textsc{StreamMA} & \cellcolor{gray!12}\textbf{62.08} & \cellcolor{gray!12}\textbf{75.83} & \cellcolor{gray!12}\textbf{58.34} & \cellcolor{gray!12}\textbf{78.12} & \cellcolor{gray!12}\textbf{15.74} & \cellcolor{gray!12}\textbf{89.50} & \cellcolor{gray!12}\textbf{94.78} & \cellcolor{gray!12}99.17 & \cellcolor{gray!12}\textbf{71.70} \\
\cmidrule{2-12}
 & \multirow{2}{*}{Graph} & Serial & 60.00 & 74.17 & 52.65 & 78.45 & 14.04 & 92.25 & 99.51 & 98.80 & 71.23 \\
 &  & \cellcolor{gray!12}\textsc{StreamMA} & \cellcolor{gray!12}\textbf{62.50} & \cellcolor{gray!12}\textbf{75.42} & \cellcolor{gray!12}\textbf{56.44} & \cellcolor{gray!12}\textbf{79.63} & \cellcolor{gray!12}\textbf{16.13} & \cellcolor{gray!12}\textbf{93.08} & \cellcolor{gray!12}\textbf{99.79} & \cellcolor{gray!12}\textbf{99.32} & \cellcolor{gray!12}\textbf{72.79} \\
\midrule
\multirow{7}{*}[-0.6em]{\rotatebox{90}{GLM-5.2}}
 & -- & Single & 40.42 & 44.17 & 32.19 & 69.02 & 7.83 & 80.08 & 70.77 & 78.05 & 52.82 \\
\cmidrule{2-12}
 & \multirow{2}{*}{Chain} & Serial & 61.25 & 65.84 & 57.20 & 77.61 & 12.13 & 85.92 & \textbf{76.34} & 97.36 & 66.71 \\
 &  & \cellcolor{gray!12}\textsc{StreamMA} & \cellcolor{gray!12}\textbf{75.42} & \cellcolor{gray!12}\textbf{78.75} & \cellcolor{gray!12}\textbf{67.05} & \cellcolor{gray!12}\textbf{79.63} & \cellcolor{gray!12}\textbf{15.18} & \cellcolor{gray!12}\textbf{89.67} & \cellcolor{gray!12}75.71 & \cellcolor{gray!12}\textbf{97.66} & \cellcolor{gray!12}\textbf{72.38} \\
\cmidrule{2-12}
 & \multirow{2}{*}{Tree} & Serial & 61.25 & 64.17 & 48.11 & 76.94 & 12.31 & 87.75 & 83.09 & 96.15 & 66.22 \\
 &  & \cellcolor{gray!12}\textsc{StreamMA} & \cellcolor{gray!12}\textbf{76.25} & \cellcolor{gray!12}\textbf{80.83} & \cellcolor{gray!12}\textbf{65.16} & \cellcolor{gray!12}\textbf{79.13} & \cellcolor{gray!12}\textbf{16.17} & \cellcolor{gray!12}\textbf{88.92} & \cellcolor{gray!12}\textbf{84.13} & \cellcolor{gray!12}\textbf{96.61} & \cellcolor{gray!12}\textbf{73.40} \\
\cmidrule{2-12}
 & \multirow{2}{*}{Graph} & Serial & 52.92 & 61.67 & 50.76 & 77.27 & 11.94 & 85.67 & 74.53 & 96.00 & 63.85 \\
 &  & \cellcolor{gray!12}\textsc{StreamMA} & \cellcolor{gray!12}\textbf{73.75} & \cellcolor{gray!12}\textbf{84.17} & \cellcolor{gray!12}\textbf{65.16} & \cellcolor{gray!12}\textbf{77.94} & \cellcolor{gray!12}\textbf{15.12} & \cellcolor{gray!12}\textbf{89.25} & \cellcolor{gray!12}\textbf{77.87} & \cellcolor{gray!12}\textbf{97.29} & \cellcolor{gray!12}\textbf{72.57} \\
\bottomrule
\end{tabular}
}
\caption{\textbf{Effectiveness across eight benchmarks and three topologies.} \textsc{StreamMA} (gray) vs.\ Single and Serial on Claude Opus~4.6-High, GPT-5.4-None, and GLM-5.2-None. Avg.\ (\%): unweighted mean; \textbf{bold}: higher in each \{Topology, Method\} pair. Single: one row per backbone, no topology axis. Each cell averages 3 runs (8 on AIME\,2025/26 and HMMT\,2026 due to small test sets). \textsc{StreamMA} outperforms both baselines in every Avg.\ cell.}
\label{tab:main_results}
\end{table*}

\subsection{Experimental Setup}
\label{sec:setup}
\noindent\textbf{Benchmarks. }
To assess robustness, we evaluate on eight benchmarks spanning competition mathematics (AIME\,2025, AIME\,2026, HMMT\,2026), graduate-level science (GPQA-Diamond~\citep{gpqa}, HLE~\citep{DBLP:journals/corr/abs-2501-14249}), and program understanding (LiveCodeBench~\citep{livecodebench}, with sub-tasks LCB-G: code generation, LCB-E: code execution, LCB-T: test output). Evaluation is conducted via OpenCompass~\citep{opencompass}; all configuration files are provided in App.~\ref{sec:config_files}.

\noindent\textbf{Baselines. }
We compare against two baselines: (1) Single, a single agent given only the query; and (2) Serial, a pipeline where each agent waits for its predecessor to finish. \textsc{StreamMA} inherits Serial's prompts and decoding, differing only in transmission granularity: Serial transmits the full output, Stream transmits each step on completion.

\noindent\textbf{Topologies. }
Unless otherwise specified, we evaluate four-agent DAG topologies ($A^0, A^1, A^2, A^3$): Chain ($A^0 \to A^1 \to A^2 \to A^3$); Tree ($A^0 \to \{A^1, A^2\} \to A^3$, with $A^1, A^2$ in parallel); and Graph (Chain plus an extra edge $A^0 \to A^2$). These three cover the canonical DAG primitives (linear, branching, and shortcut), giving a built-in sweep of connectivity rather than a single design choice.

\noindent\textbf{Implementation Details. }
We experiment on Claude Opus~4.6-High with thinking enabled, GPT-5.4-None and GLM-5.2-None without, following each provider's API defaults; later references inherit these settings unless otherwise noted. Our LLM-as-judge evaluations use GPT-5.4-None. System prompts are provided in App.~\ref{sec:prompts}. All numbers are averaged over 3 runs per \{backbone, topology\} cell, increased to 8 on AIME\,2025/26 and HMMT\,2026 due to their smaller test sets.

\subsection{Effectiveness Analysis}
\label{sec:effectiveness_exp}

\subsubsection{Quantitative Comparison}
\label{sec:quant_comparison}

\textsc{StreamMA} outperforms Serial in every Avg.\ cell across all backbones (Tab.~\ref{tab:main_results}), whether thinking is on or off, larger gains when on. \textsc{StreamMA} only appends a one-line \texttt{END\_STEP} boundary to each downstream agent's system prompt and a short solver preamble for the root agent; all other prompt content, decoding hyperparameters, and runtime settings remain identical to Serial, ruling out extra prompt engineering as the source of improvement. On Claude Opus~4.6, \textsc{StreamMA} lifts Avg.\ accuracy by $+7.3$\,pp averaged over the three topologies (peak $+22.4$\,pp on HMMT\,2026, Chain).

\textsc{StreamMA}'s gain over Serial scales inversely with Serial's strength. The three topologies span a wide Serial baseline on Claude (Avg.\ $72.92$--$79.43$), giving a built-in sweep of Serial strength rather than a topology comparison: at Serial Avg.\ $79.43$ (Tree) the gain is $+3.38$, while at $73.48$ (Chain) and $72.92$ (Graph) it grows to $+8.22$ and $+10.42$. The same trend appears within a single benchmark on HMMT\,2026, where Serial baselines of $75.00$\,/\,$63.26$\,/\,$61.75$ (Tree\,/\,Chain\,/\,Graph) correspond to gains of $+7.20$\,/\,$+22.35$\,/\,$+20.83$. Near the ceiling (e.g., GPT-5.4 LCB-T at ${\sim}99$), cells move by less than $0.6$\,pp. \refthm{thm:1}{1} captures this: \textsc{StreamMA}'s gain comes from weakening the impact of Serial's step-level errors, so a Serial run with high $p_j$ leaves few errors to weaken, while one with low $p_j$ leaves many and gains more. GPT-5.4 and GLM-5.2 exhibit the same trend, confirming the pattern is backbone- and topology-agnostic.

\subsubsection{Case Study}
\label{sec:case_study_exp}

\begin{figure*}[t]
\definecolor{serialbar}{HTML}{BDD3E8}
\definecolor{streambar}{HTML}{F2D2A8}
\definecolor{serialnum}{HTML}{7C6A85}
\definecolor{streamnum}{HTML}{6E55A0}
\begin{minipage}[t]{0.48\textwidth}
\centering
\begin{tcolorbox}[colback=teal!3!white,colframe=teal!60!black,arc=2pt,boxrule=0.4pt,left=4pt,right=4pt,top=3pt,bottom=3pt,boxsep=1pt,width=\linewidth]
\centering
{\small\textbf{Case Study:}\;\;Stream\,{\color{green!55!black}$\checkmark$}\;\;Single\,{\color{red!65!black}$\times$}\;\;Serial\,{\color{red!65!black}$\times$}}\\[2pt]
{\footnotesize\textit{Setup:}\;\;$\mathrm{Agent}^{1}\!\to\!\mathrm{Agent}^{2}$ chain,\;$S{=}8$ steps}\\[1pt]
\tcbline
{\footnotesize\textbf{(a) Step-level correctness of \boldmath$\mathrm{Agent}^{1}$}}\\[1pt]
\begin{tikzpicture}[xscale=0.95,yscale=0.47,>=stealth]
  \draw[->](0,0)--(4.50,0)node[right,font=\scriptsize]{$j$};
  \foreach \xi/\s in {0.325/1,0.825/2,1.325/3,1.825/4,2.325/5,2.825/6,3.325/7,3.825/8}
    {\draw[gray!50](\xi,0.05)--(\xi,-0.05); \node[font=\tiny,gray!60]at(\xi,-0.32){\s};}
  \draw[->](0,0)--(0,2.5)node[above,font=\scriptsize]{$p_j$};
  \foreach \yi/\yl in {0/0,2.20/1}
    {\draw[gray!45](0.05,\yi)--(-0.05,\yi); \node[font=\tiny,gray!65]at(-0.28,\yi){\yl};}
  \draw[teal!55!black,thick]
    (0.325,2.200)--(0.825,2.200)--(1.325,0.000)--(1.825,0.000)--(2.325,0.000)--(2.825,0.000)--(3.325,0.000)--(3.825,0.000);
  \foreach \x/\y in {0.325/2.200,0.825/2.200,1.325/0.000,1.825/0.000,2.325/0.000,2.825/0.000,3.325/0.000,3.825/0.000}
    \filldraw[teal!55!black](\x,\y)circle(1.3pt);
  \foreach \x in {0.325,0.825}
    \node[yshift=7pt,font=\fontsize{5}{5}\selectfont\bfseries,green!55!black] at (\x,2.200) {$\checkmark$};
  \foreach \x in {1.325,1.825,2.325,2.825,3.325,3.825}
    \node[yshift=7pt,font=\fontsize{5}{5}\selectfont\bfseries,red!65!black] at (\x,0.000) {$\times$};
\end{tikzpicture}
\tcbline
{\footnotesize\textbf{(b) Analysis via \refthm{thm:1}{1}}}\\[1pt]
{\footnotesize$\bar{p}\!\triangleq\!\tfrac{1}{S}\!\sum_{j=1}^{S}\!p_j \,{=}\, 0.25$}\\[4pt]
{\footnotesize$p_{\mathrm{head}}\!\triangleq\!\tfrac{2}{S(S+1)}\!\sum_{j=1}^{S}\!(S{+}1{-}j)\,p_j \,{\approx}\, 0.42$}\\[4pt]
{\footnotesize$p_{\mathrm{tail}}\!\triangleq\!\tfrac{2}{S(S-1)}\!\sum_{j=2}^{S}\!(j{-}1)\,p_j \,{\approx}\, 0.04$}\\[5pt]
{\small$\begin{aligned}
&\Rightarrow\;p_{\mathrm{tail}}\text{ is very low},\;p_{\mathrm{head}}\!\gg\!p_{\mathrm{tail}} \\
&\Rightarrow\;\text{head-strong, tail-weak shape} \\
&\Rightarrow\;\text{Stream advantage (\refthm{thm:1}{1})}
\end{aligned}$}
\end{tcolorbox}
\captionof{figure}{\textbf{Case study for \refthm{thm:1}{1}.} (a) Verdicts of $\mathrm{Agent}^{1}$ (\textcolor{green!55!black}{$\checkmark$}: $p_j{=}1$; \textcolor{red!65!black}{$\times$}: $p_j{=}0$). (b) $\bar{p}/p_{\mathrm{head}}/p_{\mathrm{tail}}$ place this run in Case~I.b, the Stream-advantage regime.}
\label{fig:case_study}
\end{minipage}
\hfill
\begin{minipage}[t]{0.48\textwidth}
\centering
\begin{tcolorbox}[colback=teal!3!white,colframe=teal!60!black,arc=2pt,boxrule=0.4pt,left=4pt,right=4pt,top=4pt,bottom=4pt,boxsep=1pt,width=\linewidth]
\centering
{\small\textbf{Step-Level Perturbation}}\\[3pt]
{\footnotesize\textit{Setup:}\;\;$\mathrm{Agent}^{1}\!\to\!\mathrm{Agent}^{2}$ chain,\;$S{=}4$ steps}\\[2pt]
{\footnotesize\textit{Probe:}\;\;perturb $\mathrm{Agent}^{1}$'s steps\,(\tikz[baseline=-0.5ex]\filldraw[fill=white,draw=gray!60,line width=0.4pt](0,0)rectangle(0.26,0.26);\,clean\,/\,\tikz[baseline=-0.5ex]\fill[gray!85!black](0,0)rectangle(0.26,0.26);\,perturbed)}\\[2pt]
{\footnotesize\textit{Outcome:}\;\;$\mathrm{Agent}^{2}$ acc:\,\tikz[baseline=-0.4ex]\fill[serialbar](0,0)rectangle(0.35,0.14);\,Serial / \tikz[baseline=-0.4ex]\fill[streambar](0,0)rectangle(0.35,0.14);\,Stream}
\tcbline
\begin{tikzpicture}[>=stealth, xscale=1.05, yscale=0.78]
  \node[font=\scriptsize\bfseries, anchor=south] at (0.78, 8.80) {$\mathbf{Agent}^{1}$ steps};
  \node[font=\scriptsize\bfseries, anchor=south] at (4.20, 8.80) {$\mathbf{Agent}^{2}$ accuracy (\%)};

  \fill[teal!55!black] (-0.10, 8.30) rectangle (-0.04, 8.62);
  \node[font=\scriptsize\itshape, black, anchor=west] at (0.00, 8.45) {\refthm{thm:1}{1} Case~I --- tail perturbed};
  \fill[teal!55!black] (-0.10, 5.20) rectangle (-0.04, 5.52);
  \node[font=\scriptsize\itshape, black, anchor=west] at (0.00, 5.35) {\refthm{thm:1}{1} Case~II/III --- head perturbed};

  \filldraw[fill=white,draw=gray!60,line width=0.4pt] (0.00, 7.825) rectangle (0.35, 8.175);
  \filldraw[fill=white,draw=gray!60,line width=0.4pt] (0.40, 7.825) rectangle (0.75, 8.175);
  \filldraw[fill=white,draw=gray!60,line width=0.4pt] (0.80, 7.825) rectangle (1.15, 8.175);
  \fill[gray!85!black]  (1.20, 7.825) rectangle (1.55, 8.175);
  \fill[serialbar]      (2.45, 8.04) rectangle (4.795, 8.22);
  \fill[streambar]      (2.45, 7.78) rectangle (5.635, 7.96);
  \node[font=\tiny, black, anchor=west] at (2.50, 8.13) {$67.0$};
  \node[font=\tiny, black, anchor=west] at (2.50, 7.87) {$91.0$};
  \node[font=\tiny\bfseries, green!55!black, anchor=west] at (5.96, 8.00) {$+24.0$};

  \filldraw[fill=white,draw=gray!60,line width=0.4pt] (0.00, 6.925) rectangle (0.35, 7.275);
  \filldraw[fill=white,draw=gray!60,line width=0.4pt] (0.40, 6.925) rectangle (0.75, 7.275);
  \fill[gray!85!black]  (0.80, 6.925) rectangle (1.15, 7.275);
  \fill[gray!85!black]  (1.20, 6.925) rectangle (1.55, 7.275);
  \fill[serialbar]      (2.45, 7.14) rectangle (4.655, 7.32);
  \fill[streambar]      (2.45, 6.88) rectangle (5.320, 7.06);
  \node[font=\tiny, black, anchor=west] at (2.50, 7.23) {$63.0$};
  \node[font=\tiny, black, anchor=west] at (2.50, 6.97) {$82.0$};
  \node[font=\tiny\bfseries, green!55!black, anchor=west] at (5.96, 7.10) {$+19.0$};

  \filldraw[fill=white,draw=gray!60,line width=0.4pt] (0.00, 6.025) rectangle (0.35, 6.375);
  \fill[gray!85!black]  (0.40, 6.025) rectangle (0.75, 6.375);
  \fill[gray!85!black]  (0.80, 6.025) rectangle (1.15, 6.375);
  \fill[gray!85!black]  (1.20, 6.025) rectangle (1.55, 6.375);
  \fill[serialbar]      (2.45, 6.24) rectangle (5.110, 6.42);
  \fill[streambar]      (2.45, 5.98) rectangle (5.285, 6.16);
  \node[font=\tiny, black, anchor=west] at (2.50, 6.33) {$76.0$};
  \node[font=\tiny, black, anchor=west] at (2.50, 6.07) {$81.0$};
  \node[font=\tiny\bfseries, green!55!black, anchor=west] at (5.96, 6.20) {$+5.0$};

  \fill[gray!85!black]  (0.00, 4.725) rectangle (0.35, 5.075);
  \filldraw[fill=white,draw=gray!60,line width=0.4pt] (0.40, 4.725) rectangle (0.75, 5.075);
  \filldraw[fill=white,draw=gray!60,line width=0.4pt] (0.80, 4.725) rectangle (1.15, 5.075);
  \filldraw[fill=white,draw=gray!60,line width=0.4pt] (1.20, 4.725) rectangle (1.55, 5.075);
  \fill[serialbar]     (2.45, 4.94) rectangle (5.845, 5.12);
  \fill[streambar]     (2.45, 4.68) rectangle (4.655, 4.86);
  \node[font=\tiny, black, anchor=west] at (2.50, 5.03) {$97.0$};
  \node[font=\tiny, black, anchor=west] at (2.50, 4.77) {$63.0$};
  \node[font=\tiny\bfseries, red!75!black, anchor=west] at (5.96, 4.90) {$-34.0$};

  \fill[gray!85!black]  (0.00, 3.825) rectangle (0.35, 4.175);
  \fill[gray!85!black]  (0.40, 3.825) rectangle (0.75, 4.175);
  \filldraw[fill=white,draw=gray!60,line width=0.4pt] (0.80, 3.825) rectangle (1.15, 4.175);
  \filldraw[fill=white,draw=gray!60,line width=0.4pt] (1.20, 3.825) rectangle (1.55, 4.175);
  \fill[serialbar]     (2.45, 4.04) rectangle (5.950, 4.22);
  \fill[streambar]     (2.45, 3.78) rectangle (4.690, 3.96);
  \node[font=\tiny, black, anchor=west] at (2.50, 4.13) {$100.0$};
  \node[font=\tiny, black, anchor=west] at (2.50, 3.87) {$64.0$};
  \node[font=\tiny\bfseries, red!75!black, anchor=west] at (5.96, 4.00) {$-36.0$};

  \fill[gray!85!black]  (0.00, 2.925) rectangle (0.35, 3.275);
  \fill[gray!85!black]  (0.40, 2.925) rectangle (0.75, 3.275);
  \fill[gray!85!black]  (0.80, 2.925) rectangle (1.15, 3.275);
  \filldraw[fill=white,draw=gray!60,line width=0.4pt] (1.20, 2.925) rectangle (1.55, 3.275);
  \fill[serialbar]     (2.45, 3.14) rectangle (5.880, 3.32);
  \fill[streambar]     (2.45, 2.88) rectangle (4.725, 3.06);
  \node[font=\tiny, black, anchor=west] at (2.50, 3.23) {$98.0$};
  \node[font=\tiny, black, anchor=west] at (2.50, 2.97) {$65.0$};
  \node[font=\tiny\bfseries, red!75!black, anchor=west] at (5.96, 3.10) {$-33.0$};
\end{tikzpicture}
\end{tcolorbox}
\captionof{figure}{\textbf{Step-Level Perturbation.} Fixing $\mathrm{Agent}^{1}$'s output, we perturb its steps and measure $\mathrm{Agent}^{2}$'s accuracy; \textcolor{green!55!black}{green}\,/\,\textcolor{red!75!black}{red} mark Stream's gain\,/\,loss over Serial.}
\label{fig:stepwise_perturb}
\end{minipage}
\end{figure*}

\noindent\textbf{Setup. }
On a GPQA-Diamond question, we run a two-agent chain ($\mathrm{Agent}^{1}{\to}\mathrm{Agent}^{2}$, $S{=}8$) that reuses the same Chain prompts as Sec.~\ref{sec:quant_comparison}, and binarize each step of $\mathrm{Agent}^{1}$ to $p_j\!\in\!\{0,1\}$ via LLM-as-judge ($1$ if correct, $0$ otherwise) for an instantiation of \refthm{thm:1}{1}. Fig.~\ref{fig:case_study} visualizes \textsc{StreamMA}'s per-step verdicts of $\mathrm{Agent}^{1}$; the question and per-step transcripts are deferred to App.~\ref{sec:case_study_detail}.

\noindent\textbf{Observations.}
Single, a single-agent baseline, returns the wrong answer. In Stream, $\mathrm{Agent}^{1}$ produces the trajectory in Fig.~\ref{fig:case_study}(a): steps~1--2 are correct, $\mathrm{Agent}^{1}$ first errors at step~3, and steps~3--8 are all incorrect. In Serial, an independent run of $\mathrm{Agent}^{1}$ produces the same failure pattern. This matches Case~I.b of \refthm{thm:1}{1}: a reliable head and an unreliable tail. Serial only sees $\mathrm{Agent}^{1}$'s full output, so $\mathrm{Agent}^{2}$ inherits the wrong tail and returns the wrong answer. Stream forwards the reliable head by step~2; by the time the unreliable tail arrives, $\mathrm{Agent}^{2}$ has already re-derived the answer from this prefix and returns the correct one. This is the Stream advantage predicted by \refthm{thm:1}{1}.

\subsubsection{Step-Level Perturbation}
\label{sec:stepwise_perturb}

\def\clns{\tikz[baseline=-0.3ex]\filldraw[fill=white,draw=gray!60,line width=0.3pt](0,0)rectangle(0.18,0.18);\kern1pt}%
\def\ptbs{\tikz[baseline=-0.3ex]\fill[gray!85!black](0,0)rectangle(0.18,0.18);\kern1pt}%
\noindent\textbf{Setup. }
For a cleaner verification of \refthm{thm:1}{1}, we construct a controlled experiment on an $\mathrm{Agent}^{1}{\to}\mathrm{Agent}^{2}$ chain with $S{=}4$ steps.
We use an LLM to construct two parallel $\mathrm{Agent}^{1}$ trajectories: a clean one deriving the gold answer and a perturbed counterpart that diverges from the clean one toward a wrong answer.
A 4-bit perturbation pattern swaps in \clns\,(clean) or \ptbs\,(perturbed) at each step, yielding a fixed $\mathrm{Agent}^{1}$ output; $\mathrm{Agent}^{2}$ is then run on it under Serial or Stream, with accuracy averaged over 100 runs per pattern.
Results on 6 patterns in Fig.~\ref{fig:stepwise_perturb}. Details in App.~\ref{sec:stepwise_perturb_detail}.

\noindent\textbf{Observations.}
The two pattern families produce opposite outcomes: Stream wins on tail-perturbed patterns (up to ${+}24.0$\,pp), Serial wins on head-perturbed patterns (down to ${-}36.0$\,pp).
Tail-perturbed patterns (\clns\clns\clns\ptbs, \clns\clns\ptbs\ptbs, \clns\ptbs\ptbs\ptbs) all fall in Case~I; Stream's advantage shrinks monotonically as the perturbed tail extends (${+}24.0{\to}{+}19.0{\to}{+}5.0$), tracking the predicted decay as $p_{\mathrm{tail}}$ drops.
Head-perturbed patterns span two regimes: \ptbs\clns\clns\clns and \ptbs\ptbs\clns\clns fall in Case~II, while the heavily poisoned \ptbs\ptbs\ptbs\clns reaches Case~III. In both, the curves are nearly flat regardless of depth: Serial holds at ${\sim}100\%$ while Stream trails by ${\sim}35$\,pp, trapped by the perturbed prefix.
Mirror pairs expose \refthm{thm:1}{1}'s head--tail asymmetry: \clns\clns\clns\ptbs vs.\ \ptbs\clns\clns\clns flips Stream's gap from ${+}24.0$ to ${-}34.0$\,pp.

\subsubsection{Role and Tool Generalization}
\label{sec:generalization}

\begin{table}[t]
\centering
\small
\setlength{\tabcolsep}{2pt}
\resizebox{\columnwidth}{!}{%
\begin{tabular}{l cc cc}
\toprule
\multirow{2}{*}{\textbf{Role}} & \multicolumn{2}{c}{\textbf{GLM-5.2}} & \multicolumn{2}{c}{\textbf{GPT-5.4}} \\
\cmidrule(lr){2-3} \cmidrule(lr){4-5}
 & Serial & \textsc{StreamMA} & Serial & \textsc{StreamMA} \\
\midrule
Adversarial Verify & 50.00 & \textbf{59.85} & 54.55 & \textbf{56.44} \\
Debate             & 57.58 & \textbf{63.26} & 56.06 & \textbf{61.36} \\
Decompose \& Merge & 40.53 & \textbf{45.08} & 50.38 & \textbf{56.82} \\
Reviewer Corrector & 59.85 & \textbf{67.80} & 51.89 & \textbf{56.44} \\
Self Consistency   & 36.36 & \textbf{40.15} & 45.45 & \textbf{46.21} \\
\bottomrule
\end{tabular}%
}
\caption{\textbf{Generalization to roles.} Accuracy (\%) across five roles with varying agent counts and topologies.}
\label{tab:role_generalization}
\end{table}

\noindent\textbf{Roles.} To verify that \textsc{StreamMA}'s gain generalizes beyond a fixed role design, we evaluate five role protocols on HMMT\,2026 with GLM-5.2 and GPT-5.4 over 8 runs each. The protocols differ in agent count, DAG structure, and role semantics:
(1)~Adversarial Verify: 3 agents, triangle DAG ($A^0{\to}A^1,\,A^0{\to}A^2,\,A^1{\to}A^2$); Solver, Attacker, Judge.
(2)~Debate: 2 debaters over two rounds, a 4-node chain ($A^0{\to}A^1{\to}A^0{\to}A^1$); proponent/opponent.
(3)~Decompose \& Merge: 4 agents, tree ($A^0{\to}\{A^1,A^2\}{\to}A^3$); decomposer, two parallel solvers, integrator.
(4)~Reviewer Corrector: 3 agents, chain ($A^0{\to}A^1{\to}A^2$); solver, reviewer, final corrector.
(5)~Self Consistency: 3 agents, tree ($\{A^0,A^1\}{\to}A^2$); two parallel solvers, adjudicator.
As Tab.~\ref{tab:role_generalization} shows, \textsc{StreamMA} improves over Serial in all ten \{role, backbone\} cells by +0.76 to +9.85\,pp, generalizing readily across diverse role assignments and DAG structures.

\begin{table}[t]
\centering
\small
\setlength{\tabcolsep}{2pt}
\resizebox{\columnwidth}{!}{%
\begin{tabular}{l c c c c}
\toprule
\textbf{Method} & \textbf{Loose\,/\,Strict (\%)}\,$\uparrow$ & \textbf{Cost (\$)}\,$\downarrow$ & \textbf{Latency (s)}\,$\downarrow$ & \textbf{Speedup}\,$\uparrow$ \\
\midrule
Serial            & 66.1\,/\,23.9 & 6.42 & 34.1 & 1.00$\times$ \\
\textsc{StreamMA} & \textbf{67.2\,/\,24.3} & \textbf{6.39} & \textbf{29.1} & \textbf{1.17}$\times$ \\
\bottomrule
\end{tabular}%
}
\caption{\textbf{Generalization to tools.} Answer quality and efficiency on FanOutQA with search-tool calls.}
\label{tab:tool_calling}
\end{table}

\noindent\textbf{Tools.} To verify that \textsc{StreamMA}'s gain generalizes to tool-augmented agents, we evaluate a Solver$\to$Reviewer chain ($A^0{\to}A^1$) with GPT-5.4 on the 310-question FanOutQA dev set~\citep{zhu2024fanoutqa}, a multi-hop factual QA benchmark where each question decomposes into a median of 6 sub-questions, each requiring an independent web search. The Solver issues one \texttt{[SEARCH]} call per sub-question and emits \texttt{END\_STEP} after each search result returns. Under \textsc{StreamMA}, each completed step is immediately forwarded to the Reviewer, which begins verification while the Solver continues searching. Serial and \textsc{StreamMA} differ only in this transmission timing, with prompts kept identical. We adopt FanOutQA's official loose and strict accuracy: loose credits the fraction of gold answers recalled, strict requires all of them.
Averaged over 3 runs and reported in Tab.~\ref{tab:tool_calling}, \textsc{StreamMA} improves over Serial on loose and strict accuracy by +1.1 and +0.4\,pp while also achieving $1.17\times$ speedup at lower cost, generalizing readily to agents that call external tools.

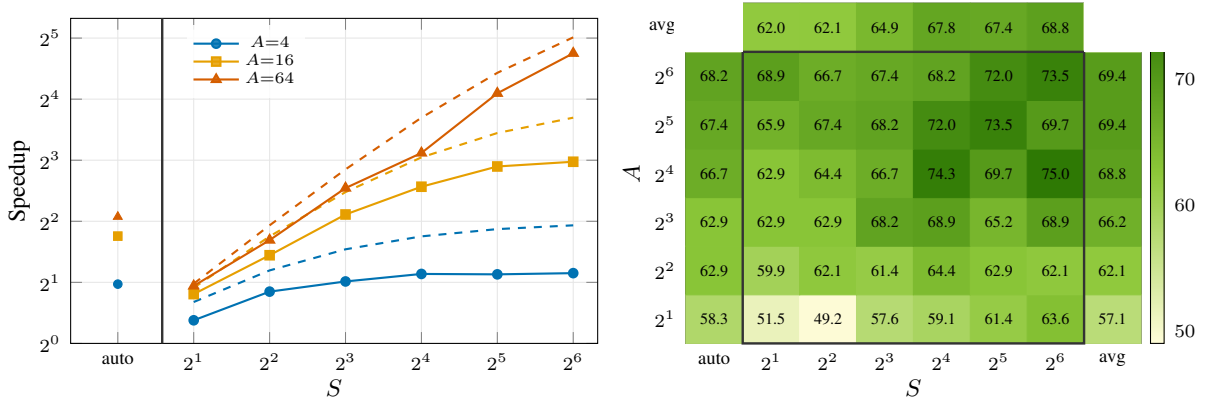
\begin{figure*}[t]
\centering
\begin{subfigure}[t]{0.495\textwidth}
\centering
\definecolor{cAfour}{HTML}{0072B2}
\definecolor{cAsixteen}{HTML}{E69F00}
\definecolor{cAsixtyfour}{HTML}{D55E00}
\begin{tikzpicture}
\begin{axis}[
  scale only axis=true,
  width=0.88\linewidth, height=4.8cm,
  xmode=log, log basis x={2}, ymode=log, log basis y={2},
  xtick={1,2,4,8,16,32,64},
  xticklabels={auto,$2^1$,$2^2$,$2^3$,$2^4$,$2^5$,$2^6$},
  xlabel={$S$}, ylabel={Speedup},
  xmin=0.65, xmax=80, ymin=1, ymax=40,
  grid=both, grid style={gray!20},
  legend style={font=\tiny, at={(0.22,0.98)}, anchor=north west,
                draw=none, fill=white, fill opacity=0.85, text opacity=1,
                row sep=-3pt, inner sep=2pt},
  legend columns=1,
  tick label style={font=\scriptsize}, label style={font=\footnotesize},
  xlabel shift=-4pt, ylabel shift=-4pt,
]
\addplot[cAfour, mark=*, mark size=1.6pt, thick]
  coordinates {(2,1.30)(4,1.80)(8,2.02)(16,2.20)(32,2.19)(64,2.22)};
\addlegendentry{$A{=}4$}
\addplot[cAfour, dashed, thick, forget plot]
  coordinates {(2,1.60)(4,2.29)(8,2.91)(16,3.37)(32,3.66)(64,3.82)};
\addplot[cAsixteen, mark=square*, mark size=1.6pt, thick]
  coordinates {(2,1.75)(4,2.72)(8,4.32)(16,5.92)(32,7.45)(64,7.86)};
\addlegendentry{$A{=}16$}
\addplot[cAsixteen, dashed, thick, forget plot]
  coordinates {(2,1.88)(4,3.37)(8,5.57)(16,8.26)(32,10.89)(64,12.96)};
\addplot[cAsixtyfour, mark=triangle*, mark size=1.9pt, thick]
  coordinates {(2,1.92)(4,3.23)(8,5.83)(16,8.68)(32,17.09)(64,26.91)};
\addlegendentry{$A{=}64$}
\addplot[cAsixtyfour, dashed, thick, forget plot]
  coordinates {(2,1.97)(4,3.82)(8,7.21)(16,12.96)(32,21.56)(64,32.25)};
\addplot[cAfour,      only marks, mark=*,         mark size=1.6pt, forget plot]
  coordinates {(1,1.96)};
\addplot[cAsixteen,   only marks, mark=square*,   mark size=1.6pt, forget plot]
  coordinates {(1,3.38)};
\addplot[cAsixtyfour, only marks, mark=triangle*, mark size=1.9pt, forget plot]
  coordinates {(1,4.21)};
\draw[black!80, line width=0.9pt] (axis cs:1.5,1) -- (axis cs:1.5,40);
\end{axis}
\end{tikzpicture}
\end{subfigure}
\hfill
\begin{subfigure}[t]{0.495\textwidth}
\centering
\pgfplotsset{
  colormap={uacc}{rgb(0cm)=(1.0,0.99,0.85); rgb(0.5cm)=(0.55,0.78,0.22); rgb(1cm)=(0.15,0.45,0.0)},
}
\begin{tikzpicture}
\begin{axis}[
  set layers,
  scale only axis=true,
  width=0.76\linewidth, height=4.8cm,
  enlargelimits=false, clip=false,
  xtick={0,1,2,3,4,5,6,7}, xticklabels={auto,$2^1$,$2^2$,$2^3$,$2^4$,$2^5$,$2^6$,avg},
  ytick={0,1,2,3,4,5,6}, yticklabels={$2^1$,$2^2$,$2^3$,$2^4$,$2^5$,$2^6$,avg},
  xlabel={$S$}, ylabel={$A$},
  tick style={draw=none},
  tick label style={font=\scriptsize}, label style={font=\footnotesize},
  xlabel shift=-4pt, ylabel shift=-4pt,
  colormap name=uacc,
  point meta min=49, point meta max=76,
  colorbar, colorbar style={width=0.18cm, font=\tiny, xshift=-0.55cm},
  nodes near coords={%
    \pgfmathtruncatemacro\isempty{(\coordindex==48)||(\coordindex==55)}%
    \ifnum\isempty=0\pgfmathprintnumber[fixed,fixed zerofill,precision=1]{\pgfplotspointmeta}\fi
  },
  every node near coord/.append style={font=\tiny, anchor=center, /pgf/number format/assume math mode=true, text=black},
]
\addplot[matrix plot*, point meta=explicit, mesh/cols=8] table[meta index=2, header=false] {
0 0 58.34
1 0 51.52
2 0 49.24
3 0 57.58
4 0 59.10
5 0 61.37
6 0 63.64
7 0 57.1
0 1 62.88
1 1 59.85
2 1 62.13
3 1 61.37
4 1 64.40
5 1 62.88
6 1 62.13
7 1 62.1
0 2 62.88
1 2 62.88
2 2 62.88
3 2 68.19
4 2 68.94
5 2 65.16
6 2 68.94
7 2 66.2
0 3 66.67
1 3 62.88
2 3 64.40
3 3 66.67
4 3 74.25
5 3 69.70
6 3 75.00
7 3 68.8
0 4 67.43
1 4 65.91
2 4 67.43
3 4 68.19
4 4 71.97
5 4 73.49
6 4 69.70
7 4 69.4
0 5 68.19
1 5 68.94
2 5 66.67
3 5 67.43
4 5 68.19
5 5 71.97
6 5 73.49
7 5 69.4
0 6 49
1 6 62.0
2 6 62.1
3 6 64.9
4 6 67.8
5 6 67.4
6 6 68.8
7 6 49
};
\begin{pgfonlayer}{axis foreground}
\fill[white, draw=none] (axis cs:-0.7,5.5) rectangle (axis cs:0.5,7.0);
\fill[white, draw=none] (axis cs:6.5,5.5) rectangle (axis cs:8.0,7.0);
\end{pgfonlayer}
\draw[black!80, line width=0.9pt] (axis cs:0.5,-0.5) rectangle (axis cs:6.5,5.5);
\end{axis}
\end{tikzpicture}
\end{subfigure}
\caption{\textbf{Step-level scaling law.} \textit{Left:} speedup scaling in $S$; measured (solid) vs.\ theoretical maximum speedup $AS/(S{+}A{-}1)$ from \refthm{thm:2}{2} (dashed). \textit{Right:} accuracy scaling in $S$, with \textit{avg} marginals (main block boxed).}
\label{fig:efficiency}
\end{figure*}

\subsection{Efficiency Analysis}

\subsubsection{Step-Level Scaling Law}
\label{sec:step_scaling_law}

\noindent\textbf{Setup. }
We adopt the Chain $\mathrm{Agent}^{1}\!\to\!\cdots\!\to\!\mathrm{Agent}^{A}$ with each agent producing $S$ reasoning steps, and sweep $A\!\in\!\{2,4,8,16,32,64\}$ and $S\!\in\!\{2,4,8,16,32,64,\mathrm{auto}\}$ on HMMT\,2026; $S{=}\mathrm{auto}$ lets $\mathrm{Agent}^{1}$ autonomously decide the step count. Each $(A,S)$ cell reports per-question averages of speedup and accuracy over 4 independent runs, where speedup is the sum of per-agent API call times divided by Stream's measured wall-clock runtime. KV-cache reuse is disabled; for visual clarity the left panel of Fig.~\ref{fig:efficiency} plots $A\!\in\!\{4,16,64\}$ only, while the full $6{\times}6$ grid appears in the right heatmap. This $AS/(S+A-1)$ form is the simplification of \refthm{thm:2}{2} under our no-KV-cache setup with decode much slower than prefill ($r_{v_{dp}}\!\to\!0$).

\noindent\textbf{Observations.}
\textbf{(i)} Measured speedup tracks the $AS/(S{+}A{-}1)$ scaling pattern across the $(A,S)$ grid (Fig.~\ref{fig:efficiency}, left), validating \refthm{thm:2}{2}; at the $A{=}64$, $S{=}64$ corner the measured speedup reaches $26.9\times$, attaining $83\%$ of the theoretical bound ($32.3\times$); the gap arises because the bound assumes $r_{v_{dp}}\!\to\!0$, while actual $r_{v_{dp}}>0$ in GPT-5.4. Notably, the $S{=}\mathrm{auto}$ markers fall well short of the large-$S$ speedups across all $A$, because LLMs do not autonomously scale up their step count and default to a coarse granularity; the large-$S$ regime must be explicitly unlocked through prompting.
\textbf{(ii)} Accuracy increases jointly with $A$ and $S$ across the heatmap (Fig.~\ref{fig:efficiency}, right). The right \textit{avg} column reproduces agent-count scaling~\citep{scaling}, while the top \textit{avg} row is near-monotone in $S$, evidencing a step-level scaling law: at fixed $A$, accuracy improves with more reasoning steps. The two scaling dimensions (agent count and step count) are complementary, not redundant: at $A{=}64$, the $S{=}\mathrm{auto}$ baseline (LLM-decided $S$) reaches $68.2\%$, and increasing $S$ to $64$ further raises accuracy to $73.5\%$, indicating that step-level scaling delivers measurable gains beyond agent-count scaling alone.

\subsubsection{Cost Analysis}
\label{sec:realworld_cost}

\begin{figure}[t]
\centering
\begin{tikzpicture}
\begin{axis}[
    name=upper,
    scale only axis=true,
    width=0.88\columnwidth, height=3.4cm,
    ylabel={Accuracy (\%)},
    xmode=log, log basis x=10, log ticks with fixed point,
    xtick={0.3, 0.5, 1, 2, 3, 6},
    xticklabels={},          
    xmin=0.2, xmax=8,
    ymin=68, ymax=94,
    ytick={70, 80, 90},
    grid=major, grid style={dashed,gray!25},
    label style={font=\footnotesize},
    tick label style={font=\scriptsize},
    ylabel shift=-4pt,
    legend style={font=\scriptsize, fill opacity=0.85, draw=none,
                  at={(0.99,0.02)}, anchor=south east,
                  xshift=1pt, yshift=-3pt,
                  row sep=-2pt, inner sep=2pt},
    legend cell align=left,
    every axis plot/.append style={line join=round},
]
\fill[red!85!black, opacity=0.15]
    (axis cs:0.3442, 78.79) -- (axis cs:1.6111, 90.91) --
    (axis cs:2.7502, 90.91) -- (axis cs:0.4661, 78.79) -- cycle;
\addplot[color=gray!50!black, mark=*, mark size=1.6pt, thick]
    coordinates { (0.3969, 70.45) (1.0311, 81.82) (5.4586, 89.39) };
\addlegendentry{Serial$\times N$}
\node[font=\scriptsize, gray!50!black, anchor=north, inner sep=1pt, yshift=-1pt] at (axis cs:0.3969, 70.45) {$1$};
\node[font=\scriptsize, gray!50!black, anchor=north, inner sep=1pt, yshift=-1pt] at (axis cs:1.0311, 81.82) {$4$};
\node[font=\scriptsize, gray!50!black, anchor=north, inner sep=1pt, yshift=-1pt] at (axis cs:5.4586, 89.39) {$16$};
\addplot[color=red!85!black, mark=triangle*, mark size=2pt, thick]
    coordinates { (0.4661, 78.79) (2.7502, 90.91) };
\addlegendentry{Stream$\times N$ ($h{=}0$)}
\node[font=\scriptsize, red!85!black, anchor=north, inner sep=1pt, yshift=-1pt] at (axis cs:0.4661, 78.79) {$1$};
\node[font=\scriptsize, red!85!black, anchor=north, inner sep=1pt, yshift=-1pt] at (axis cs:2.7502, 90.91) {$4$};
\addplot[color=red!85!black, mark=triangle*, mark size=2pt, thick, densely dashed]
    coordinates { (0.3442, 78.79) (1.6111, 90.91) };
\addlegendentry{Stream$\times N$ ($h{=}1$)}
\addlegendimage{only marks, mark=star, mark size=3pt,
    color=blue!70!black, mark options={line width=0.8pt}}
\addlegendentry{Single}
\end{axis}
\begin{axis}[
    name=lower,
    at={(upper.south west)}, anchor=north west,
    yshift=-5pt,
    scale only axis=true,
    width=0.88\columnwidth, height=1.0cm,
    xlabel={Per-question cost (USD)},
    xmode=log, log basis x=10, log ticks with fixed point,
    xtick={0.3, 0.5, 1, 2, 3, 6},
    xmin=0.2, xmax=8,
    ymin=44, ymax=52,
    ytick={48},
    grid=major, grid style={dashed,gray!25},
    label style={font=\footnotesize},
    tick label style={font=\scriptsize},
    xlabel shift=-4pt,
]
\addplot[only marks, color=blue!70!black, mark=star, mark size=3pt,
         mark options={line width=0.8pt}]
    coordinates { (0.2649, 48.11) };  
\end{axis}
\draw[white, line width=1.6pt] ([xshift=0.3pt] upper.south west) -- ([xshift=-0.3pt] upper.south east);
\draw[white, line width=1.6pt] ([xshift=0.3pt] lower.north west) -- ([xshift=-0.3pt] lower.north east);
\draw[thick] ([xshift=-1mm, yshift=2.5pt] lower.north west) -- ++(2mm,-5pt);
\draw[thick] ([xshift=0.5mm, yshift=2.5pt] lower.north west) -- ++(2mm,-5pt);
\draw[thick] ([xshift=-1.5mm, yshift=2.5pt] lower.north east) -- ++(2mm,-5pt);
\draw[thick] ([xshift=0mm, yshift=2.5pt] lower.north east) -- ++(2mm,-5pt);
\end{tikzpicture}
\caption{\textbf{Cost--accuracy Pareto frontiers.} Each frontier tracks accuracy vs.\ cost as $N\!\in\!\{1,4,16\}$ chain replicas run in parallel and majority-vote on the final answer; larger $N$ trades higher compute for higher accuracy. Red shaded: KV-cache hit rate $h\!\in\!(0,1)$, bounded by Stream$\times N$ at $h{=}0$ (solid) and $h{=}1$ (dashed).}
\label{fig:cost_pareto}
\end{figure}
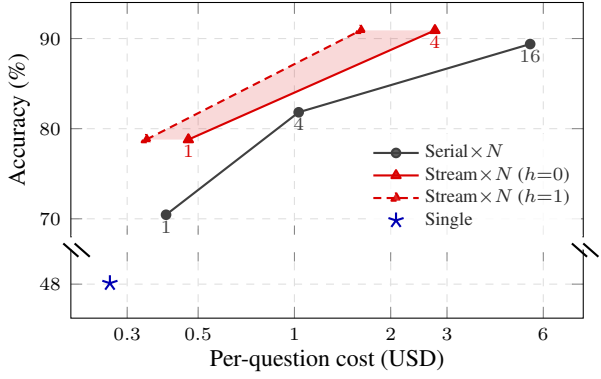

\noindent\textbf{Setup. }
We adopt the Chain $\mathrm{Agent}^{1}\!\to\!\mathrm{Agent}^{2}\!\to\!\mathrm{Agent}^{3}$ with each agent producing $S{=}3$ reasoning steps (prompts adapted from Sec.~\ref{sec:effectiveness_exp}), and sweep both Stream$\times N$ and Serial$\times N$ over $N\!\in\!\{1,4,16\}$ parallel chain replicas with majority voting on HMMT\,2026 with Claude Opus~4.6 (API pricing: $\$5$\,/\,$\$25$\,/\,$\$0.5$ per million input\,/\,output\,/\,cache-read tokens). Each $N$ point reports per-question averages of cost and accuracy over 4 independent runs. Let $h\!\in\![0,1]$ denote the KV-cache hit rate.

\noindent\textbf{Observations.}
\textbf{(i)}~The Stream$\times N$ frontier strictly Pareto-dominates Serial$\times N$: Stream$\times 4$ ($\$2.75$, $90.9\%$) achieves higher accuracy than Serial$\times 16$ ($\$5.46$, $89.4\%$) at half the cost, despite Serial running $4{\times}$ more replicas, jointly confirming Stream's accuracy advantage (\refthm{thm:1}{1}) and cost advantage (\refthm{thm:3}{3}).
\textbf{(ii)}~The shaded KV-cache band ($h\!\in\!(0,1)$) compresses each Stream$\times N$ point leftward by up to ${\sim}1.7\times$ with no accuracy loss (e.g., Stream$\times 4$ drops from $\$2.75$ to $\$1.61$ at fixed $90.9\%$); this $h{=}0\!\to\!h{=}1$ compression ratio is governed by the cache-to-prefill price ratio $r_{c_{cp}}$ in \refthm{thm:3}{3}, and further widens the Pareto gap.
\textbf{(iii)}~Without voting ($N{=}1$), Stream at full cache ($\$0.34$, $78.8\%$) strictly dominates Serial ($\$0.40$, $70.5\%$): the Pareto advantage stems from the protocol itself, not from voting amplification.

%% file: latex/secs/06_conclusion.tex
\section{Conclusion}

We introduced \textsc{StreamMA}, a reasoning-step-level streaming multi-agent system
for LLM reasoning that replaces the prevailing ``generate-then-transfer''
paradigm with immediate per-step forwarding.
The central finding is counter-intuitive: streaming reduces latency
and improves effectiveness, because the non-uniform quality structure of
multi-step LLM reasoning makes when context arrives matter more than
how much context arrives.
We formalized this mechanism through three closed-form theorems covering effectiveness
ordering, speedup upper bound, and cost ratio, all empirically validated across eight
benchmarks, three frontier LLMs, and three topologies.
Beyond the protocol, we uncovered a ``step-level scaling law'':
a dimension orthogonal to agent-count scaling that monotonically improves
effectiveness and speedup, suggesting that the design space of multi-agent reasoning
systems is richer than recognized.

%% file: latex/secs/07_limitation.tex
\section{Limitations}

\textsc{StreamMA} pipelines reasoning at the granularity of steps, and thus applies to tasks whose solution admits step decomposition. Modern reasoning workloads (mathematics, code, science) satisfy this property under chain-of-thought prompting, the de facto standard. Tasks whose output is atomic (e.g., single-token classification) offer no intermediate steps to stream and fall outside this regime; this reflects the task's structure rather than a constraint of \textsc{StreamMA}, and equally bounds any step-based reasoning paradigm (e.g., chain-of-thought).

Furthermore, the stream protocol is not universally optimal across all step-correctness profiles. \refthm{thm:1}{1} characterizes six regimes: stream execution is strictly preferable only when agents exhibit head-strong, tail-weak correctness patterns relative to the breakeven $p^*$; in the remaining regimes, serial or single execution may match or exceed it. Rather than weakening \textsc{StreamMA}, this boundary is precisely predicted by our theoretical framework. \refthm{thm:1}{1} therefore serves as a principled protocol selector: given the step-correctness profile and the breakeven $p^*$ of a target task, practitioners can directly consult the theorem to determine which execution protocol best suits their workload.

%% file: latex/secs/08_appendix.tex
\section{Appendix}

\begin{tcolorbox}[colback=green!3!white, colframe=green!45!black, arc=3pt, boxrule=0.6pt,
  title={\small\textbf{Appendix Contents}}, fonttitle=\small,
  left=6pt, right=6pt, top=4pt, bottom=5pt]
\footnotesize
\setlength{\parindent}{0pt}%
\setlength{\parfillskip}{0pt}%
\setlength{\rightskip}{0pt plus 1fil}%
\setlength{\parskip}{2pt}%
\newcommand{\apptocone}[1]{\vspace{1pt}\hangindent=2.6em\hangafter=1%
  \hyperref[#1]{\makebox[2.6em][l]{\textbf{\ref*{#1}}}\textbf{\nameref*{#1}}}\hfill\textbf{\pageref{#1}}\par}
\newcommand{\apptoctwo}[2]{\hangindent=4.2em\hangafter=1%
  \hspace{2.6em}\hyperref[#2]{#1}\dotfill\pageref{#2}\par}
\apptocone{sec:notation}
\apptocone{sec:detailed-effectiveness-analysis}
\apptoctwo{Proof of Core Identities}{sec:proof-core}
\apptoctwo{Proof of Theorem 1}{sec:proof-t1}
\apptoctwo{Discussion of Practical Scenarios}{sec:discussion}
\apptocone{sec:detailed-speedup-analysis}
\apptoctwo{Latency of Serial Protocol}{sec:lat-serial}
\apptoctwo{Latency of Stream Protocol}{sec:lat-stream}
\apptoctwo{Proof of Theorem 2}{sec:proof-t2}
\apptoctwo{Numerical Example: Speedup}{sec:num-speedup}
\apptocone{sec:detailed-cost-analysis}
\apptoctwo{Cost of Serial Protocol}{sec:cost-serial}
\apptoctwo{Cost of Stream Protocol}{sec:cost-stream}
\apptoctwo{Proof of Theorem 3}{sec:proof-t3}
\apptoctwo{Numerical Example: Cost}{sec:num-cost}
\apptocone{sec:case_study_detail}
\apptocone{sec:stepwise_perturb_detail}
\apptocone{sec:config_files}
\apptocone{sec:stream-dag-extension}
\apptocone{sec:prompts}
\apptocone{sec:ai_assistants}
\apptocone{sec:potential_risks}
\end{tcolorbox}
\onecolumn

\subsection{Notation Summary}
\label{sec:notation}

Tab.~\ref{tab:notation} summarizes the notation used in the theoretical analysis, covering parameters for the effectiveness and efficiency (speedup and cost) analyses. For brevity, derivations in the appendix use a chain topology as a running example; the generalization to arbitrary DAGs is discussed at the end of each proof section.

\begin{table}[ht]
\centering
\footnotesize
\renewcommand{\arraystretch}{1.25}
\caption{Notation summary.}
\label{tab:notation}
\begin{tabular}{ll}
\toprule
\textbf{Symbol} & \textbf{Description} \\
\midrule
\multicolumn{2}{l}{\textit{General parameters}} \\
$A$ & Number of agents \\
$S$ & Number of reasoning steps \\
$o_s^a$ & Number of output tokens of $\mathrm{Agent}^a$ at step $s$ \\
$O_a$ & Total number of output tokens of $\mathrm{Agent}^a$; $O_a \triangleq \sum_{s=1}^{S} o_s^a$ \\
$\bar{O}$ & Per-agent per-step output token count, averaged across agents; $\bar{O} \triangleq \frac{1}{A}\sum_{a=1}^{A} \frac{O_a}{S}$ \\
$P_a$ & Number of tokens in system prompt of $\mathrm{Agent}^a$ \\
$\bar{P}$ & Average system prompt length; $\bar{P} \triangleq \frac{1}{A}\sum_{a=1}^{A} P_a$ \\
$O_{\Sigma}$ & Total number of output tokens across all agents; $O_{\Sigma} \triangleq \sum_{a=1}^{A} O_a$ \\
\midrule
\multicolumn{2}{l}{\textit{Effectiveness analysis} ($\mathrm{Agent}^a \to \mathrm{Agent}^{a+1}$)} \\
$\mathrm{sCorr}^{\mathrm{mode}}$ & Mean step-level correctness, $\mathrm{mode}\!\in\!\{\mathrm{stream},\mathrm{serial},\mathrm{single}\}$ \\
$p_j$ & Correctness probability of step $j$; $p_j \in [0,1]$ \\
$\delta$ & Expected downstream step-correctness gain from a correct upstream step in context \\
$\varepsilon$ & Expected downstream step-correctness drop from an incorrect upstream step in context \\
$\mu_j$ & Expected step-correctness change from upstream step $j$; $\mu_j \triangleq p_j\delta - (1-p_j)\varepsilon$ \\
$p^{*}$ & Minimum step correctness for context to be beneficial; $p^* \triangleq \varepsilon/(\delta+\varepsilon)$ \\
$\bar{p}$ & Mean step correctness; $\bar{p} \triangleq \frac{1}{S}\sum_{j=1}^S p_j$ \\
$p_{\mathrm{tail}}$ & Tail-weighted mean step correctness; $p_{\mathrm{tail}} \triangleq \frac{2}{S(S-1)}\sum_{j=2}^{S}(j-1)\,p_j$ \\
$p_{\mathrm{head}}$ & Head-weighted mean step correctness; $p_{\mathrm{head}} \triangleq \frac{2}{S(S+1)}\sum_{j=1}^{S}(S+1-j)\,p_j$ \\
\midrule
\multicolumn{2}{l}{\textit{Speedup analysis}} \\
$C_s^a$ & Total number of context tokens in the $s$-th call of $\mathrm{Agent}^a$ \\
$h_s^a$ & KV-cache hit rate of $\mathrm{Agent}^a$'s $s$-th call; $h_s^a\in[0,1]$ \\
$\alpha$ & Average number of non-cached prefill tokens per output token; $\alpha \triangleq \frac{\sum_{a=1}^{A}\sum_{s=1}^{S} (1-h_s^a)\,C_s^a}{AS\bar{O}}$ \\
$\beta$ & Average number of cache-hit tokens per output token; $\beta \triangleq \frac{\sum_{a=1}^{A}\sum_{s=1}^{S} h_s^a\,C_s^a}{AS\bar{O}}$ \\
$v_d,\,v_p,\,v_c$ & Decode / prefill / cache-read speed (tok/s); $v_c \gg v_p \gg v_d$ \\
$r_{v_{dp}}$ & Decode-to-prefill speed ratio; $r_{v_{dp}} \triangleq v_d/v_p$ \\
$r_{v_{dc}}$ & Decode-to-cache-read speed ratio; $r_{v_{dc}} \triangleq v_d/v_c$ \\
$r_{po}$ & Prompt-to-per-step-output ratio; $r_{po} \triangleq \bar{P}/\bar{O}$ \\
\midrule
\multicolumn{2}{l}{\textit{Cost analysis}} \\
$c_d,\,c_p,\,c_c$ & Per-token price: decode / prefill / cache; $c_d \gg c_p \gg c_c$ \\
$r_{c_{cp}}$ & Cache-to-prefill price ratio; $r_{c_{cp}} \triangleq c_c/c_p$ \\
$r_{c_{pd}}$ & Prefill-to-decode price ratio; $r_{c_{pd}} \triangleq c_p/c_d$ \\
$\rho$ & Ratio of total output token counts: Stream vs.\ Serial; $\rho \triangleq O_{\Sigma}^{\mathrm{stream}}/O_{\Sigma}^{\mathrm{serial}}$ \\
\bottomrule
\end{tabular}
\end{table}

\subsection{Detailed Effectiveness Analysis}
\label{sec:detailed-effectiveness-analysis}

Let $\mathrm{sCorr}^{\mathrm{mode}}$, $\mathrm{mode}\in\{\mathrm{serial},\mathrm{stream},\mathrm{single}\}$, denote the mean step-level expected correctness of the downstream agent. \textit{This is a theoretical quantity positively correlated with task-level accuracy: a higher fraction of correct reasoning steps makes a correct final answer more likely.}
\begin{tcolorbox}[colback=green!4!white, colframe=green!45!black, arc=3pt, boxrule=0.6pt]
\label{thm:1:app}\textbf{Theorem 1} (Effectiveness Ordering). \textit{Depending on how $\bar{p}$, $p_{\mathrm{head}}$, $p_{\mathrm{tail}}$ compare to $p^*$, the sCorr ordering among the three modes falls into six cases:}
\smallskip

\noindent
\begin{tabular}{@{\hspace{0.5em}}l@{\quad}l@{}}
  \multicolumn{2}{@{\hspace{0.5em}}l}{\textbf{(I) Stream advantage} ~[$p_{\mathrm{head}} > p^*$ and $p_{\mathrm{tail}} < p^*$]\textbf{:}} \\[2pt]
  \quad\textit{(a)~If $\bar{p} > p^*$:} & $\mathrm{sCorr}^{\mathrm{stream}} > \mathrm{sCorr}^{\mathrm{serial}} > \mathrm{sCorr}^{\mathrm{single}}$ \\[3pt]
  \quad\textit{(b)~If $\bar{p} < p^*$:} & $\mathrm{sCorr}^{\mathrm{stream}} > \mathrm{sCorr}^{\mathrm{single}} > \mathrm{sCorr}^{\mathrm{serial}}$ \\[5.2pt]
  \multicolumn{2}{@{\hspace{0.5em}}l}{\textbf{(II) Serial advantage} ~[$\bar{p} > p^*$ and $p_{\mathrm{tail}} > p^*$]\textbf{:}} \\[2pt]
  \quad\textit{(a)~If $p_{\mathrm{head}} > p^*$:}    & $\mathrm{sCorr}^{\mathrm{serial}} > \mathrm{sCorr}^{\mathrm{stream}} > \mathrm{sCorr}^{\mathrm{single}}$ \\[3pt]
  \quad\textit{(b)~If $p_{\mathrm{head}} < p^*$:}    & $\mathrm{sCorr}^{\mathrm{serial}} > \mathrm{sCorr}^{\mathrm{single}} > \mathrm{sCorr}^{\mathrm{stream}}$ \\[5.2pt]
  \multicolumn{2}{@{\hspace{0.5em}}l}{\textbf{(III) Single advantage} ~[$p_{\mathrm{head}} < p^*$ and $\bar{p} < p^*$]\textbf{:}} \\[2pt]
  \quad\textit{(a)~If $p_{\mathrm{tail}} < p^*$:} & $\mathrm{sCorr}^{\mathrm{single}} > \mathrm{sCorr}^{\mathrm{stream}} > \mathrm{sCorr}^{\mathrm{serial}}$ \\[3pt]
  \quad\textit{(b)~If $p_{\mathrm{tail}} > p^*$:} & $\mathrm{sCorr}^{\mathrm{single}} > \mathrm{sCorr}^{\mathrm{serial}} > \mathrm{sCorr}^{\mathrm{stream}}$ \\
\end{tabular}
\end{tcolorbox}

\subsubsection{Proof of Core Identities}
\label{sec:proof-core}
Consider adjacent agents $\mathrm{Agent}^a$ (upstream) and $\mathrm{Agent}^{a+1}$ (downstream) in the chain: the upstream produces $S$-step reasoning $A^a_1, \ldots, A^a_S$; the downstream produces outputs $A^{a+1}_1, \ldots, A^{a+1}_S$. The three modes differ only in which upstream steps are visible to the downstream when generating $A^{a+1}_s$:

\begin{center}
\begin{tabular}{cc}
\toprule
Mode & Condition when predicting $A^{a+1}_s$ \\
\midrule
Single & $\mathbb{P}(A^{a+1}_s \mid A^{a+1}_{<s})$ \\
Stream & $\mathbb{P}(A^{a+1}_s \mid A^{a+1}_{<s},\, A^a_{\leq s})$ \\
Serial & $\mathbb{P}(A^{a+1}_s \mid A^{a+1}_{<s},\, A^a_{1:S})$ \\
\bottomrule
\end{tabular}
\end{center}

\noindent where $\mathbb{P}(A^{a+1}_s \mid \cdot)$ is the expected step-level correctness of $A^{a+1}_s$, abbreviated $\mathbb{P}^{\mathrm{mode}}(A^{a+1}_s)$ for each row.

Let $\Delta_j$ denote the change in downstream step-level correctness when step $j$ is added to a context containing steps $1,\ldots,j{-}1$. We parametrize $\Delta_j$ by conditional expectations: $\delta \triangleq \mathbb{E}[\Delta_j \mid A^a_j\text{ correct},\, A^a_{<j}\text{ in context}]>0$ is the expected gain when step $j$ is correct, and $\varepsilon \triangleq \mathbb{E}[-\Delta_j \mid A^a_j\text{ incorrect},\, A^a_{<j}\text{ in context}]>0$ the expected loss when step $j$ is incorrect. \textit{We assume $\delta, \varepsilon$ are step-position-independent for simplicity; in general, $(\delta, \varepsilon)$ generalize to vectors $(\delta_j, \varepsilon_j)$ and the same derivation applies.} Since step $j$ is correct with probability $p_j$, the expected step-correctness change is
$$\mu_j \triangleq p_j \cdot \delta - (1-p_j)\cdot\varepsilon = p_j(\delta+\varepsilon)-\varepsilon.$$
The breakeven threshold $p^* \triangleq \varepsilon/(\delta+\varepsilon)$ satisfies $\mu_j > 0$ if and only if $p_j > p^*$. With position-dependent $\delta_j,\varepsilon_j$, the threshold becomes $\varepsilon_j/(\delta_j+\varepsilon_j)$ and may vary by step. The three modes differ in upstream context seen: Single sees none, Stream sees a prefix, Serial sees all $S$ steps. With Single as the baseline, summing the corresponding $\mu_j$ gives the step-level correctness gains of Stream and Serial:

$$\mathbb{P}^{\mathrm{stream}}(A^{a+1}_s) - \mathbb{P}^{\mathrm{single}}(A^{a+1}_s) = \sum_{j=1}^s \mu_j, \qquad \mathbb{P}^{\mathrm{serial}}(A^{a+1}_s) - \mathbb{P}^{\mathrm{single}}(A^{a+1}_s) = \sum_{j=1}^S \mu_j$$

By definition, $\mathrm{sCorr}^{\mathrm{mode}} \triangleq \frac{1}{S}\sum_{s=1}^S \mathbb{P}^{\mathrm{mode}}(A^{a+1}_s)$ averages step-level correctness over $s$. Applying the same average to the step-level differences above yields the sCorr gaps between modes:

\begin{equation}
\mathrm{sCorr}^{\mathrm{stream}} - \mathrm{sCorr}^{\mathrm{single}} = \frac{1}{S}\sum_{s=1}^{S}\sum_{j=1}^{s}\mu_j = \frac{1}{S}\sum_{j=1}^{S}(S-j+1)\,\mu_j \tag{a}\label{eq:identity-a}
\end{equation}

\begin{equation}
\mathrm{sCorr}^{\mathrm{serial}} - \mathrm{sCorr}^{\mathrm{single}} = \frac{1}{S}\sum_{s=1}^{S}\sum_{j=1}^{S}\mu_j = \sum_{j=1}^{S}\mu_j \tag{b}\label{eq:identity-b}
\end{equation}

Subtracting \eqref{eq:identity-a} from \eqref{eq:identity-b}:

\begin{equation}
\mathrm{sCorr}^{\mathrm{serial}} - \mathrm{sCorr}^{\mathrm{stream}} = \frac{1}{S}\sum_{j=1}^S (j-1)\,\mu_j \tag{c}\label{eq:identity-c}
\end{equation}

The signs of \eqref{eq:identity-a}--\eqref{eq:identity-c} follow from how the head ($p_{\mathrm{head}}$), mean ($\bar{p}$), and tail ($p_{\mathrm{tail}}$) per-step correctness compare to $p^*$ (Tab.~\ref{tab:notation}), fixing the ordering among Single, Stream, and Serial and establishing \refthm{thm:1:app}{1}.

\noindent\textbf{Remark} (Boundary cases). Equality in any condition yields the corresponding equality in the ordering.

\subsubsection{Proof of Theorem 1}
\label{sec:proof-t1}

\noindent We prove each case by inverting identities \eqref{eq:identity-a}--\eqref{eq:identity-c}: given a target sCorr ordering, the signs of \eqref{eq:identity-a}--\eqref{eq:identity-c} determine conditions on $p_{\mathrm{head}}$, $\bar{p}$, and $p_{\mathrm{tail}}$ relative to $p^*$. These variables are not prescribed \emph{a priori} but derived from the proof below; their parametric definitions (Tab.~\ref{tab:notation}) absorb the resulting coefficients.

\noindent\textbf{Case I.a} ($\mathrm{sCorr}^{\mathrm{stream}} > \mathrm{sCorr}^{\mathrm{serial}} > \mathrm{sCorr}^{\mathrm{single}}$)\textbf{.}
This ordering requires $\eqref{eq:identity-a}>0$, $\eqref{eq:identity-b}>0$, $\eqref{eq:identity-c}<0$.
\begin{itemize}
  \item \textit{Expanding \eqref{eq:identity-b}:}
  \[ \sum_{j=1}^S \mu_j = \sum_{j=1}^S \bigl[(\delta+\varepsilon)p_j - \varepsilon\bigr] = S\bigl[(\delta+\varepsilon)\bar{p} - \varepsilon\bigr], \]
  which is positive if and only if $\bar{p} > p^*$.
  \item \textit{Expanding \eqref{eq:identity-c}:} Since the weight at $j=1$ vanishes,
  \begin{align*}
  \frac{1}{S}\sum_{j=2}^S (j-1)\,\mu_j &= \frac{1}{S}\sum_{j=2}^S (j-1)\bigl[(\delta+\varepsilon)p_j - \varepsilon\bigr] \\
      &= \frac{1}{S}\!\left[(\delta+\varepsilon)\sum_{j=2}^S(j-1)p_j - \varepsilon\sum_{j=2}^S(j-1)\right].
  \end{align*}
  Since $\sum_{j=2}^S(j-1)=\tfrac{S(S-1)}{2}$ and $\sum_{j=2}^S(j-1)p_j = \tfrac{S(S-1)}{2}p_{\mathrm{tail}}$ by definition:
  \[ \eqref{eq:identity-c} = \frac{S-1}{2}\bigl[(\delta+\varepsilon)p_{\mathrm{tail}} - \varepsilon\bigr], \]
  which is negative if and only if $p_{\mathrm{tail}} < p^*$.
  \item \textit{Sign of \eqref{eq:identity-a}:} Since $\eqref{eq:identity-b} > 0$ and $\eqref{eq:identity-c} < 0$, we have $\eqref{eq:identity-a} = \eqref{eq:identity-b} - \eqref{eq:identity-c} > 0$, so $\mathrm{sCorr}^{\mathrm{stream}} > \mathrm{sCorr}^{\mathrm{serial}} > \mathrm{sCorr}^{\mathrm{single}}$ holds when $\bar{p} > p^*$ and $p_{\mathrm{tail}} < p^*$, implying $p_{\mathrm{head}} > p^*$.
\end{itemize}

\noindent\textbf{Case I.b} ($\mathrm{sCorr}^{\mathrm{stream}} > \mathrm{sCorr}^{\mathrm{single}} > \mathrm{sCorr}^{\mathrm{serial}}$)\textbf{.}
This ordering requires $\eqref{eq:identity-a}>0$, $\eqref{eq:identity-b}<0$, $\eqref{eq:identity-c}<0$.
\begin{itemize}
  \item \textit{Expanding \eqref{eq:identity-b}:} By the same expansion as in Case~I.a,
  \[ \eqref{eq:identity-b} = \sum_{j=1}^S \mu_j = S\bigl[(\delta+\varepsilon)\bar{p} - \varepsilon\bigr], \]
  which is negative if and only if $\bar{p} < p^*$.
  \item \textit{Expanding \eqref{eq:identity-a}:} Multiplying both sides by $S$ and substituting $\mu_j = (\delta+\varepsilon)p_j - \varepsilon$:
  \begin{align*}
  S\!\cdot\!\eqref{eq:identity-a}
  &= \sum_{j=1}^{S}(S+1-j)\bigl[(\delta+\varepsilon)p_j - \varepsilon\bigr] \\
  &= (\delta+\varepsilon)\sum_{j=1}^{S}(S+1-j)\,p_j - \varepsilon\sum_{j=1}^{S}(S+1-j).
  \end{align*}
  Substituting $\sum_{j=1}^{S}(S+1-j) = \tfrac{S(S+1)}{2}$ and $\sum_{j=1}^{S}(S+1-j)\,p_j = \tfrac{S(S+1)}{2}p_{\mathrm{head}}$ by definition:
  \[ \eqref{eq:identity-a} = \tfrac{S+1}{2}\bigl[(\delta+\varepsilon)p_{\mathrm{head}}-\varepsilon\bigr], \]
  which is positive if and only if $p_{\mathrm{head}} > p^*$.
  \item \textit{Sign of \eqref{eq:identity-c}:} Since $\eqref{eq:identity-c} = \eqref{eq:identity-b} - \eqref{eq:identity-a}$, $\eqref{eq:identity-b} < 0$ and $\eqref{eq:identity-a} > 0$ give $\eqref{eq:identity-c} < 0$, so $\mathrm{sCorr}^{\mathrm{stream}} > \mathrm{sCorr}^{\mathrm{single}} > \mathrm{sCorr}^{\mathrm{serial}}$ holds when $p_{\mathrm{head}} > p^*$ and $\bar{p} < p^*$, implying $p_{\mathrm{tail}} < p^*$.
\end{itemize}

\noindent\textbf{Case II.a} ($\mathrm{sCorr}^{\mathrm{serial}} > \mathrm{sCorr}^{\mathrm{stream}} > \mathrm{sCorr}^{\mathrm{single}}$)\textbf{.}
This ordering requires $\eqref{eq:identity-a}>0$, $\eqref{eq:identity-b}>0$, $\eqref{eq:identity-c}>0$.
\begin{itemize}
  \item \textit{Expanding \eqref{eq:identity-a}:} By the same expansion as in Case~I.b,
  \[ \eqref{eq:identity-a} = \tfrac{S+1}{2}\bigl[(\delta+\varepsilon)p_{\mathrm{head}}-\varepsilon\bigr], \]
  which is positive if and only if $p_{\mathrm{head}} > p^*$.
  \item \textit{Expanding \eqref{eq:identity-c}:} By the same expansion as in Case~I.a,
  \[ \eqref{eq:identity-c} = \tfrac{S-1}{2}\bigl[(\delta+\varepsilon)p_{\mathrm{tail}}-\varepsilon\bigr], \]
  which is positive if and only if $p_{\mathrm{tail}} > p^*$.
  \item \textit{Sign of \eqref{eq:identity-b}:} Since $\eqref{eq:identity-b} = \eqref{eq:identity-a} + \eqref{eq:identity-c}$, both positive, $\eqref{eq:identity-b} > 0$, so $\mathrm{sCorr}^{\mathrm{serial}} > \mathrm{sCorr}^{\mathrm{stream}} > \mathrm{sCorr}^{\mathrm{single}}$ holds when $p_{\mathrm{head}} > p^*$ and $p_{\mathrm{tail}} > p^*$, implying $\bar{p} > p^*$.
\end{itemize}

\noindent\textbf{Case II.b} ($\mathrm{sCorr}^{\mathrm{serial}} > \mathrm{sCorr}^{\mathrm{single}} > \mathrm{sCorr}^{\mathrm{stream}}$)\textbf{.}
This ordering requires $\eqref{eq:identity-a}<0$, $\eqref{eq:identity-b}>0$, $\eqref{eq:identity-c}>0$.
\begin{itemize}
  \item \textit{Expanding \eqref{eq:identity-b}:} By the same expansion as in Case~I.a,
  \[ \sum_{j=1}^S \mu_j = S\bigl[(\delta+\varepsilon)\bar{p}-\varepsilon\bigr], \]
  which is positive if and only if $\bar{p} > p^*$.
  \item \textit{Expanding \eqref{eq:identity-a}:} By the same expansion as in Case~I.b,
  \[ \eqref{eq:identity-a} = \tfrac{S+1}{2}\bigl[(\delta+\varepsilon)p_{\mathrm{head}}-\varepsilon\bigr], \]
  which is negative if and only if $p_{\mathrm{head}} < p^*$.
  \item \textit{Sign of \eqref{eq:identity-c}:} Since $\eqref{eq:identity-c} = \eqref{eq:identity-b} - \eqref{eq:identity-a}$, $\eqref{eq:identity-b} > 0$ and $\eqref{eq:identity-a} < 0$ give $\eqref{eq:identity-c} > 0$, so $\mathrm{sCorr}^{\mathrm{serial}} > \mathrm{sCorr}^{\mathrm{single}} > \mathrm{sCorr}^{\mathrm{stream}}$ holds when $\bar{p} > p^*$ and $p_{\mathrm{head}} < p^*$, implying $p_{\mathrm{tail}} > p^*$.
\end{itemize}

\noindent\textbf{Case III.a} ($\mathrm{sCorr}^{\mathrm{single}} > \mathrm{sCorr}^{\mathrm{stream}} > \mathrm{sCorr}^{\mathrm{serial}}$)\textbf{.}
This ordering requires $\eqref{eq:identity-a}<0$, $\eqref{eq:identity-b}<0$, $\eqref{eq:identity-c}<0$.
\begin{itemize}
  \item \textit{Expanding \eqref{eq:identity-a}:} By the same expansion as in Case~I.b,
  \[ \eqref{eq:identity-a} = \tfrac{S+1}{2}\bigl[(\delta+\varepsilon)p_{\mathrm{head}}-\varepsilon\bigr], \]
  which is negative if and only if $p_{\mathrm{head}} < p^*$.
  \item \textit{Expanding \eqref{eq:identity-c}:} By the same expansion as in Case~I.a,
  \[ \eqref{eq:identity-c} = \tfrac{S-1}{2}\bigl[(\delta+\varepsilon)p_{\mathrm{tail}}-\varepsilon\bigr], \]
  which is negative if and only if $p_{\mathrm{tail}} < p^*$.
  \item \textit{Sign of \eqref{eq:identity-b}:} Since $\eqref{eq:identity-b} = \eqref{eq:identity-a} + \eqref{eq:identity-c}$, both negative, $\eqref{eq:identity-b} < 0$, so $\mathrm{sCorr}^{\mathrm{single}} > \mathrm{sCorr}^{\mathrm{stream}} > \mathrm{sCorr}^{\mathrm{serial}}$ holds when $p_{\mathrm{head}} < p^*$ and $p_{\mathrm{tail}} < p^*$, implying $\bar{p} < p^*$.
\end{itemize}

\noindent\textbf{Case III.b} ($\mathrm{sCorr}^{\mathrm{single}} > \mathrm{sCorr}^{\mathrm{serial}} > \mathrm{sCorr}^{\mathrm{stream}}$)\textbf{.}
This ordering requires $\eqref{eq:identity-a}<0$, $\eqref{eq:identity-b}<0$, $\eqref{eq:identity-c}>0$.
\begin{itemize}
  \item \textit{Expanding \eqref{eq:identity-b}:} By the same expansion as in Case~I.a,
  \[ \sum_{j=1}^S \mu_j = S\bigl[(\delta+\varepsilon)\bar{p}-\varepsilon\bigr], \]
  which is negative if and only if $\bar{p} < p^*$.
  \item \textit{Expanding \eqref{eq:identity-c}:} By the same expansion as in Case~I.a,
  \[ \eqref{eq:identity-c} = \tfrac{S-1}{2}\bigl[(\delta+\varepsilon)p_{\mathrm{tail}}-\varepsilon\bigr], \]
  which is positive if and only if $p_{\mathrm{tail}} > p^*$.
  \item \textit{Sign of \eqref{eq:identity-a}:} Since $\eqref{eq:identity-a} = \eqref{eq:identity-b} - \eqref{eq:identity-c}$, $\eqref{eq:identity-b} < 0$ and $\eqref{eq:identity-c} > 0$ give $\eqref{eq:identity-a} < 0$, so $\mathrm{sCorr}^{\mathrm{single}} > \mathrm{sCorr}^{\mathrm{serial}} > \mathrm{sCorr}^{\mathrm{stream}}$ holds when $\bar{p} < p^*$ and $p_{\mathrm{tail}} > p^*$, implying $p_{\mathrm{head}} < p^*$.
\end{itemize}

\noindent\textbf{Remark} (Extension to DAG topologies). The proof above considers a single agent pair for clarity. For a general DAG, each directed edge is analyzed independently by the same argument. When a downstream agent has multiple predecessors, the contributions from all upstream edges are additive by linearity of expectation, so the ordering determined by each edge's own $p_{\mathrm{head}}$, $\bar{p}$, $p_{\mathrm{tail}}$ composes linearly; \refthm{thm:1:app}{1} then applies per edge. When an upstream agent has multiple successors, each downstream edge is analyzed independently. Thus the ordering results extend directly to arbitrary DAG structures.

\subsubsection{Discussion of Practical Scenarios}
\label{sec:discussion}

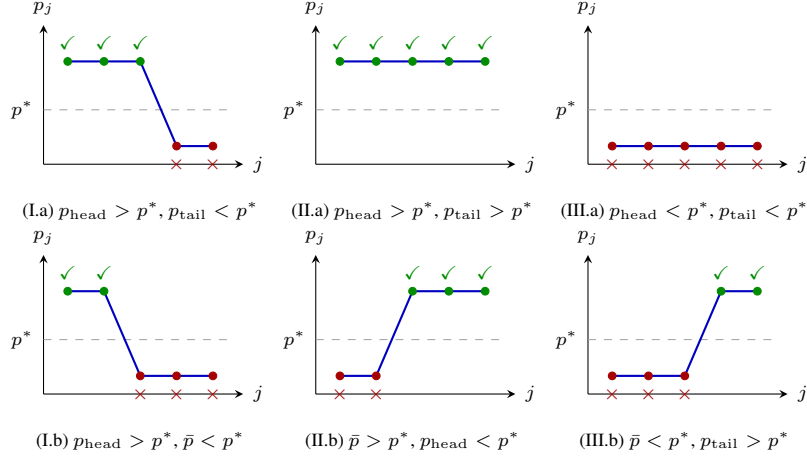
\begin{figure}[t]
\centering
\begin{tikzpicture}[scale=0.80, >=stealth]

\begin{scope}[xshift=0cm]
  \draw[->] (0,0) -- (3.3,0) node[right,font=\scriptsize] {$j$};
  \draw[->] (0,0) -- (0,2.3) node[above,font=\scriptsize] {$p_j$};
  \draw[dashed,gray!70] (0,0.9) -- (3.1,0.9);
  \node[left,font=\scriptsize] at (0,0.9) {$p^*$};
  \draw[blue!75!black,thick] (0.4,1.7)--(1.0,1.7)--(1.6,1.7)--(2.2,0.3)--(2.8,0.3);
  \foreach \x in {0.4,1.0,1.6} \filldraw[green!55!black] (\x,1.7) circle (1.8pt);
  \foreach \x in {2.2,2.8} \filldraw[red!65!black] (\x,0.3) circle (1.8pt);
  \foreach \x in {0.4,1.0,1.6} \node[font=\scriptsize,green!55!black,yshift=7pt] at (\x,1.7) {$\checkmark$};
  \foreach \x in {2.2,2.8} \node[font=\scriptsize,red!65!black,yshift=-7pt] at (\x,0.3) {$\times$};
  \node[below,font=\scriptsize] at (1.6,-0.45) {(I.a)~$p_{\mathrm{head}}>p^*$,~$p_{\mathrm{tail}}<p^*$};
\end{scope}

\begin{scope}[xshift=4.5cm]
  \draw[->] (0,0) -- (3.3,0) node[right,font=\scriptsize] {$j$};
  \draw[->] (0,0) -- (0,2.3) node[above,font=\scriptsize] {$p_j$};
  \draw[dashed,gray!70] (0,0.9) -- (3.1,0.9);
  \node[left,font=\scriptsize] at (0,0.9) {$p^*$};
  \draw[blue!75!black,thick] (0.4,1.7)--(1.0,1.7)--(1.6,1.7)--(2.2,1.7)--(2.8,1.7);
  \foreach \x in {0.4,1.0,1.6,2.2,2.8} \filldraw[green!55!black] (\x,1.7) circle (1.8pt);
  \foreach \x in {0.4,1.0,1.6,2.2,2.8} \node[font=\scriptsize,green!55!black,yshift=7pt] at (\x,1.7) {$\checkmark$};
  \node[below,font=\scriptsize] at (1.6,-0.45) {(II.a)~$p_{\mathrm{head}}>p^*$,~$p_{\mathrm{tail}}>p^*$};
\end{scope}

\begin{scope}[shift={(0cm,-3.8cm)}]
  \draw[->] (0,0) -- (3.3,0) node[right,font=\scriptsize] {$j$};
  \draw[->] (0,0) -- (0,2.3) node[above,font=\scriptsize] {$p_j$};
  \draw[dashed,gray!70] (0,0.9) -- (3.1,0.9);
  \node[left,font=\scriptsize] at (0,0.9) {$p^*$};
  \draw[blue!75!black,thick] (0.4,1.7)--(1.0,1.7)--(1.6,0.3)--(2.2,0.3)--(2.8,0.3);
  \foreach \x in {0.4,1.0} \filldraw[green!55!black] (\x,1.7) circle (1.8pt);
  \foreach \x in {1.6,2.2,2.8} \filldraw[red!65!black] (\x,0.3) circle (1.8pt);
  \foreach \x in {0.4,1.0} \node[font=\scriptsize,green!55!black,yshift=7pt] at (\x,1.7) {$\checkmark$};
  \foreach \x in {1.6,2.2,2.8} \node[font=\scriptsize,red!65!black,yshift=-7pt] at (\x,0.3) {$\times$};
  \node[below,font=\scriptsize] at (1.6,-0.45) {(I.b)~$p_{\mathrm{head}}>p^*$,~$\bar{p}<p^*$};
\end{scope}

\begin{scope}[xshift=9.0cm]
  \draw[->] (0,0) -- (3.3,0) node[right,font=\scriptsize] {$j$};
  \draw[->] (0,0) -- (0,2.3) node[above,font=\scriptsize] {$p_j$};
  \draw[dashed,gray!70] (0,0.9) -- (3.1,0.9);
  \node[left,font=\scriptsize] at (0,0.9) {$p^*$};
  \draw[blue!75!black,thick] (0.4,0.3)--(1.0,0.3)--(1.6,0.3)--(2.2,0.3)--(2.8,0.3);
  \foreach \x in {0.4,1.0,1.6,2.2,2.8} \filldraw[red!65!black] (\x,0.3) circle (1.8pt);
  \foreach \x in {0.4,1.0,1.6,2.2,2.8} \node[font=\scriptsize,red!65!black,yshift=-7pt] at (\x,0.3) {$\times$};
  \node[below,font=\scriptsize] at (1.6,-0.45) {(III.a)~$p_{\mathrm{head}}<p^*$,~$p_{\mathrm{tail}}<p^*$};
\end{scope}

\begin{scope}[shift={(4.5cm,-3.8cm)}]
  \draw[->] (0,0) -- (3.3,0) node[right,font=\scriptsize] {$j$};
  \draw[->] (0,0) -- (0,2.3) node[above,font=\scriptsize] {$p_j$};
  \draw[dashed,gray!70] (0,0.9) -- (3.1,0.9);
  \node[left,font=\scriptsize] at (0,0.9) {$p^*$};
  \draw[blue!75!black,thick] (0.4,0.3)--(1.0,0.3)--(1.6,1.7)--(2.2,1.7)--(2.8,1.7);
  \foreach \x in {0.4,1.0} \filldraw[red!65!black] (\x,0.3) circle (1.8pt);
  \foreach \x in {1.6,2.2,2.8} \filldraw[green!55!black] (\x,1.7) circle (1.8pt);
  \foreach \x in {0.4,1.0} \node[font=\scriptsize,red!65!black,yshift=-7pt] at (\x,0.3) {$\times$};
  \foreach \x in {1.6,2.2,2.8} \node[font=\scriptsize,green!55!black,yshift=7pt] at (\x,1.7) {$\checkmark$};
  \node[below,font=\scriptsize] at (1.6,-0.45) {(II.b)~$\bar{p}>p^*$,~$p_{\mathrm{head}}<p^*$};
\end{scope}

\begin{scope}[shift={(9.0cm,-3.8cm)}]
  \draw[->] (0,0) -- (3.3,0) node[right,font=\scriptsize] {$j$};
  \draw[->] (0,0) -- (0,2.3) node[above,font=\scriptsize] {$p_j$};
  \draw[dashed,gray!70] (0,0.9) -- (3.1,0.9);
  \node[left,font=\scriptsize] at (0,0.9) {$p^*$};
  \draw[blue!75!black,thick] (0.4,0.3)--(1.0,0.3)--(1.6,0.3)--(2.2,1.7)--(2.8,1.7);
  \foreach \x in {0.4,1.0,1.6} \filldraw[red!65!black] (\x,0.3) circle (1.8pt);
  \foreach \x in {2.2,2.8} \filldraw[green!55!black] (\x,1.7) circle (1.8pt);
  \foreach \x in {0.4,1.0,1.6} \node[font=\scriptsize,red!65!black,yshift=-7pt] at (\x,0.3) {$\times$};
  \foreach \x in {2.2,2.8} \node[font=\scriptsize,green!55!black,yshift=7pt] at (\x,1.7) {$\checkmark$};
  \node[below,font=\scriptsize] at (1.6,-0.45) {(III.b)~$\bar{p}<p^*$,~$p_{\mathrm{tail}}>p^*$};
\end{scope}

\end{tikzpicture}
\caption{Six canonical step-correctness profiles $p_j$, $1\le j\le S$ (solid lines) relative to the breakeven threshold $p^*$ (dashed line), corresponding to the six cases of Theorem~1, organized into three advantage regimes (columns). Left column (Stream-advantage): Case~I.a (top), Case~I.b (bottom); middle column (Serial-advantage): Case~II.a (top), Case~II.b (bottom); right column (Single-advantage): Case~III.a (top), Case~III.b (bottom).}
\label{fig:scenarios:all}
\end{figure}

\refthm{thm:1:app}{1} identifies six step-correctness regimes illustrated in Fig.~\ref{fig:scenarios:all}. The six examples fix $p^*$ for simplicity. ``Strong'' and ``weak'' describe $p_j$ relative to this threshold, which may vary by step.

\smallskip\noindent\textit{(I) Stream advantage} [$p_{\mathrm{head}} > p^*$ and $p_{\mathrm{tail}} < p^*$].

\noindent\textbf{Case I.a} ($\bar{p}>p^*$; Fig.~\ref{fig:scenarios:all} panel~I.a). The profile starts high but late steps fall below $p^*$. Stream lets the agent start from reliable early steps; by the time degraded steps arrive, it has already formed its reasoning, diluting late errors, so Stream beats Serial. Since $\bar{p}>p^*$, full context helps, so Serial beats Single.

\noindent\textbf{Case I.b} ($\bar{p}<p^*$; Fig.~\ref{fig:scenarios:all} panel~I.b). Early steps are reliable ($p_{\mathrm{head}}>p^*$), but $\bar{p}<p^*$ means later steps drag the profile below $p^*$. Stream lets the agent start from the good early steps, beating Single. Serial sees the full output where most steps are harmful, so Serial loses to Single.

\smallskip\noindent\textit{(II) Serial advantage} [$\bar{p} > p^*$ and $p_{\mathrm{tail}} > p^*$].

\noindent\textbf{Case II.a} ($p_{\mathrm{head}}>p^*$; Fig.~\ref{fig:scenarios:all} panel~II.a). Since $p_{\mathrm{head}}>p^*$ and $p_{\mathrm{tail}}>p^*$, every step helps. Serial sees all $S$ steps and benefits most, so Serial beats Stream. Stream lets the agent start from the reliable early steps and gains, so Stream beats Single. Single receives no upstream context and ranks last.

\noindent\textbf{Case II.b} ($p_{\mathrm{head}}<p^*$; Fig.~\ref{fig:scenarios:all} panel~II.b). Early steps fall below $p^*$, but $\bar{p}>p^*$ means late steps are reliable. Serial sees all steps and benefits, beating Single. Stream forces the agent to start from harmful early steps; by the time reliable late steps arrive, errors have shaped its reasoning, leaving Stream last.

\smallskip\noindent\textit{(III) Single advantage} [$p_{\mathrm{head}} < p^*$ and $\bar{p} < p^*$].

\noindent\textbf{Case III.a} ($p_{\mathrm{tail}}<p^*$; Fig.~\ref{fig:scenarios:all} panel~III.a). All steps fall below $p^*$, so upstream context is harmful. Single receives no upstream context and dominates. Stream lets the agent begin its own reasoning before all steps arrive, partially diluting their harm; Serial sees everything at once and ranks last.

\noindent\textbf{Case III.b} ($p_{\mathrm{tail}}>p^*$; Fig.~\ref{fig:scenarios:all} panel~III.b). Late steps are reliable, but $\bar{p}<p^*$ means harmful early steps drag the profile below $p^*$. Without upstream context, Single dominates. Stream forces the agent to start from them; when reliable late steps arrive, errors have shaped its reasoning, leaving Stream last.

Among the six cases, Cases~I.a and~I.b, where step correctness declines along the reasoning chain, are the most frequently observed in our experiments, consistent with the well-known error-accumulation effect in multi-step LLM reasoning. Case~II.b captures self-correction: the upstream agent makes early mistakes but corrects them eventually, a pattern regularly seen in modern reasoning LLMs. Cases~III.a and~III.b are rare: III.a corresponds to problems so hard that all steps fall below $p^*$, while III.b demands a heavily corrupted prefix yet too few remaining steps to lift $p_{\mathrm{tail}}$ above $p^*$.

\subsection{Detailed Speedup Analysis}
\label{sec:detailed-speedup-analysis}

\begin{tcolorbox}[colback=green!4!white, colframe=green!45!black, arc=3pt, boxrule=0.6pt]
\label{thm:2:app}\textbf{Theorem 2} (Speedup Upper Bound). \textit{The latency speedup of stream protocol over serial protocol is upper-bounded by:}
\[
\mathrm{Speedup} = \frac{A\bigl[(S+r_{po})\,r_{v_{dp}} + S\bigr]}{(S+A-1)(1 + \alpha\,r_{v_{dp}} + \beta\,r_{v_{dc}})}
\]
\end{tcolorbox}

\subsubsection{Latency of Serial Protocol}
\label{sec:lat-serial}

In the serial protocol, each agent issues one API call, passing its full output without KV-cache reuse.

\noindent\textbf{Root agent.} $\mathrm{Agent}^1$ prefills system prompt $P_1$ and query $O_0$, then decodes $O_1$.

\noindent\textbf{Non-root agents.} $\mathrm{Agent}^a$ ($a \geq 2$) prefills system prompt $P_a$ and predecessor output $O_{a-1}$, decodes $O_a$:

\begin{equation*}
T_{\mathrm{serial}} = \sum_{a=1}^A \left[\frac{P_a + O_{a-1}}{v_p} + \frac{O_a}{v_d}\right] \tag{1}\label{eq:serial-latency}
\end{equation*}

\subsubsection{Latency of Stream Protocol}
\label{sec:lat-stream}

\noindent\textbf{Root agent.} $\mathrm{Agent}^1$ issues a single API call: prefilling system prompt $P_1$ and query $O_0$, decoding $O_1$:
\begin{equation*}
T_1 = \frac{P_1 + O_0}{v_p} + \frac{O_1}{v_d} \tag{2}
\end{equation*}

\noindent\textbf{Non-root agents.} $\mathrm{Agent}^a$ ($a \geq 2$) issues $S$ calls. At step $s$, the context length is $C_s^a$, the KV-cache hit rate is $h_s^a$, and the output length is $o_s^a$, giving a per-step latency of:
\begin{equation*}
t_s^a = \frac{h_s^a\,C_s^a}{v_c} + \frac{(1-h_s^a)\,C_s^a}{v_p} + \frac{o_s^a}{v_d} \tag{3}\label{eq:step-latency}
\end{equation*}
Summing Eq.~\ref{eq:step-latency} over $S$ steps, the total latency of agent $a$ is:
\begin{equation*}
T_a^{\mathrm{stream}} = \sum_{s=1}^{S} t_s^a = \frac{\sum_{s=1}^{S} h_s^a\,C_s^a}{v_c} + \frac{\sum_{s=1}^{S} (1-h_s^a)\,C_s^a}{v_p} + \frac{O_a}{v_d} \tag{4}\label{eq:agent-stream-latency}
\end{equation*}

\noindent The $A$ agents form a pipeline: while $\mathrm{Agent}^a$ processes step $s$, $\mathrm{Agent}^{a-1}$ can already process step $s+1$. Let $F_s^a$ denote the wall-clock time at which $\mathrm{Agent}^a$ completes step $s$. With boundary conditions $F_0^a = F_s^0 = 0$, the exact finish time follows the recurrence:
\begin{equation*}
F_s^a = \max\!\left(F_{s-1}^a,\; F_s^{a-1}\right) + t_s^a, \quad T_{\mathrm{stream}} = F_S^A \tag{5}\label{eq:pipeline-recurrence}
\end{equation*}

\subsubsection{Proof of Theorem 2}
\label{sec:proof-t2}

\noindent Since Eq.~\ref{eq:pipeline-recurrence} admits no closed form, we lower-bound $T_{\mathrm{stream}}$, yielding an upper bound on $\mathrm{Speedup}$.

\noindent\textbf{Upper-bound derivation.} Assuming comparable workloads across agents:

\textit{Simplifying $T_{\mathrm{serial}}$.} Substituting $P_a \approx \bar{P}$ and $O_a \approx S\bar{O}$ into Eq.~\ref{eq:serial-latency}:
\begin{equation*}
\begin{aligned}
T_{\mathrm{serial}}
&= \sum_{a=1}^A \left[\frac{P_a + O_{a-1}}{v_p} + \frac{O_a}{v_d}\right]
\approx \sum_{a=1}^A \left[\frac{\bar{P} + S\bar{O}}{v_p} + \frac{S\bar{O}}{v_d}\right] \\
&= \frac{A(\bar{P} + S\bar{O})}{v_p} + \frac{AS\bar{O}}{v_d}
\end{aligned}
\end{equation*}

\textit{Simplifying $T_{\mathrm{stream}}$.} The pipeline has two phases: (1) the first step traverses all $A$ agents serially, contributing $\frac{1}{S}\sum_{a=1}^{A} T_a^{\mathrm{stream}}$; (2) each of the remaining $S-1$ steps waits only for the bottleneck agent, contributing $\frac{S-1}{S}\max_{a=1}^{A} T_a^{\mathrm{stream}}$. Hence:
\begin{equation*}
T_{\mathrm{stream}} \approx \frac{1}{S}\sum_{a=1}^{A} T_a^{\mathrm{stream}} + \frac{S-1}{S}\max_{a=1}^{A} T_a^{\mathrm{stream}} \tag{6}\label{eq:pipeline-approx}
\end{equation*}
Substituting Eq.~\ref{eq:agent-stream-latency} into Eq.~\ref{eq:pipeline-approx}. Since $O_a \approx S\bar{O}$, the decode term exits $\max_{a=1}^{A}$; replacing $\max_{a=1}^{A}$ with $\frac{1}{A}\sum_{a=1}^{A}$ (since $\max \geq \mathrm{mean}$, this lower-bounds $T_{\mathrm{stream}}$):
\begin{equation*}
\begin{aligned}
T_{\mathrm{stream}}
&\approx \frac{1}{S}\sum_{a=1}^{A}\sum_{s=1}^{S}\!\left[\frac{h_s^a C_s^a}{v_c} + \frac{(1-h_s^a)C_s^a}{v_p} + \frac{o_s^a}{v_d}\right] \\
&\quad + \frac{S-1}{S}\max_{a=1}^{A}\!\left[\sum_{s=1}^{S}\!\left(\frac{h_s^a C_s^a}{v_c} + \frac{(1-h_s^a)C_s^a}{v_p} + \frac{o_s^a}{v_d}\right)\right] \\[5.2pt]
&\approx \frac{1}{S}\sum_{a=1}^{A}\sum_{s=1}^{S}\!\left[\frac{h_s^a C_s^a}{v_c} + \frac{(1-h_s^a)C_s^a}{v_p}\right] + \frac{A\bar{O}}{v_d} \\
&\quad + \frac{S-1}{S}\max_{a=1}^{A}\!\left[\frac{\sum_{s=1}^{S} h_s^a C_s^a}{v_c} + \frac{\sum_{s=1}^{S}(1-h_s^a)C_s^a}{v_p}\right] + \frac{(S-1)\bar{O}}{v_d} \\[5.2pt]
&\approx \frac{1}{S}\sum_{a=1}^{A}\sum_{s=1}^{S}\!\left[\frac{h_s^a C_s^a}{v_c} + \frac{(1-h_s^a)C_s^a}{v_p}\right] + \frac{A\bar{O}}{v_d} \\
&\quad + \frac{S-1}{SA}\sum_{a=1}^{A}\sum_{s=1}^{S}\!\left[\frac{h_s^a C_s^a}{v_c} + \frac{(1-h_s^a)C_s^a}{v_p}\right] + \frac{(S-1)\bar{O}}{v_d} \\[5.2pt]
&= \frac{S+A-1}{SA}\sum_{a=1}^{A}\sum_{s=1}^{S}\!\left[\frac{h_s^a C_s^a}{v_c} + \frac{(1-h_s^a)C_s^a}{v_p}\right] + \frac{(A+S-1)\bar{O}}{v_d}
\end{aligned}
\tag{7}\label{eq:Tstream-merged}
\end{equation*}

Let $\alpha \triangleq \dfrac{\sum_{a=1}^{A}\sum_{s=1}^{S}(1-h_s^a)C_s^a}{AS\bar{O}}$ and $\beta \triangleq \dfrac{\sum_{a=1}^{A}\sum_{s=1}^{S}h_s^a C_s^a}{AS\bar{O}}$. Here $\alpha$ is the average uncached context tokens per output token, and $\beta$ is the cached counterpart. $C_s^a$ takes different forms: for the root agent ($a=1$), there is only one call with $C_1^1 = \bar{P}+S\bar{O}$ and $C_s^1=0$ for $s\geq 2$; for non-root agents ($a\geq 2$), at step $s$ the context accumulates as $C_s^a = \bar{P}+(2s-1)\bar{O}$ (system prompt, $s$ upstream steps, and $s-1$ own previous steps). To instantiate $\alpha$ and $\beta$, we consider two practical scenarios:
\begin{itemize}
  \item \textit{Without prefix caching} ($h_s^a=0$ for all non-root agents and steps): all context tokens require prefill, so they contribute entirely to $\alpha$:
  \begin{align*}
  \alpha &= \frac{C_1^1 + \sum_{a=2}^{A}\sum_{s=1}^{S}C_s^a}{AS\bar{O}}
       = \frac{(\bar{P}+S\bar{O}) + (A-1)\sum_{s=1}^{S}[\bar{P}+(2s-1)\bar{O}]}{AS\bar{O}} \\
       &= \frac{(\bar{P}+S\bar{O}) + (A-1)(S\bar{P}+S^2\bar{O})}{AS\bar{O}}
       \quad\bigl(\text{using } \textstyle\sum_{s=1}^{S}(2s-1)=S^2\bigr) \\
       &= \frac{\bar{O}(r_{po}+S) + (A-1)S\bar{O}(r_{po}+S)}{AS\bar{O}}
       = \frac{(r_{po}+S)\bigl[1+(A-1)S\bigr]}{AS} \\[5.2pt]
  \beta &= 0
  \end{align*}
  \item \textit{With prefix caching} ($h_s^1=0$, $h_s^a=1$ for $a\geq 2$): the root agent's call is uncached since the original query is not pre-cached, while non-root agents benefit from full prefix caching since the upstream output is already cached. Only $a=1$ contributes to $\alpha$:
  \begin{align*}
  \alpha &= \frac{C_1^1}{AS\bar{O}}
       = \frac{\bar{P}+S\bar{O}}{AS\bar{O}}
       = \frac{r_{po}+S}{AS} \\[5.2pt]
  \beta &= \frac{\sum_{a=2}^{A}\sum_{s=1}^{S}C_s^a}{AS\bar{O}}
       = \frac{(A-1)\sum_{s=1}^{S}[\bar{P}+(2s-1)\bar{O}]}{AS\bar{O}} \\
       &= \frac{(A-1)(S\bar{P}+S^2\bar{O})}{AS\bar{O}}
       = \frac{(A-1)(r_{po}+S)}{A}
  \end{align*}
\end{itemize}

Substituting $\sum_{a=1}^{A}\sum_{s=1}^{S}h_s^a C_s^a = \beta AS\bar{O}$ and $\sum_{a=1}^{A}\sum_{s=1}^{S}(1-h_s^a)C_s^a = \alpha AS\bar{O}$ into Eq.~\ref{eq:Tstream-merged}:
\begin{equation*}
\begin{aligned}
T_{\mathrm{stream}}
&\approx \frac{S+A-1}{SA}\!\left(\frac{\beta AS\bar{O}}{v_c} + \frac{\alpha AS\bar{O}}{v_p}\right) + \frac{(S+A-1)\bar{O}}{v_d} \\
&= (S+A-1)\!\left(\frac{\beta\bar{O}}{v_c} + \frac{\alpha\bar{O}}{v_p} + \frac{\bar{O}}{v_d}\right)
\end{aligned}
\end{equation*}

\textit{Speedup.} Taking $\mathrm{Speedup} = T_{\mathrm{serial}}/T_{\mathrm{stream}}$, letting $r_{po} \triangleq \bar{P}/\bar{O}$, and canceling $\bar{O}$:
\begin{equation*}
\begin{aligned}
\mathrm{Speedup}
&\approx \frac{\dfrac{A(\bar{P}+S\bar{O})}{v_p} + \dfrac{AS\bar{O}}{v_d}}{(S+A-1)\!\left(\dfrac{\beta\bar{O}}{v_c} + \dfrac{\alpha\bar{O}}{v_p} + \dfrac{\bar{O}}{v_d}\right)} \\[5.2pt]
&= \frac{A\!\left(\dfrac{S+r_{po}}{v_p} + \dfrac{S}{v_d}\right)}{(S+A-1)\!\left(\dfrac{\beta}{v_c} + \dfrac{\alpha}{v_p} + \dfrac{1}{v_d}\right)}
\end{aligned}
\end{equation*}
Multiplying numerator and denominator by $v_d$ and letting $r_{v_{dp}} \triangleq v_d/v_p$, $r_{v_{dc}} \triangleq v_d/v_c$:
\begin{equation*}
\boxed{\mathrm{Speedup} = \frac{A\!\left[(S+r_{po})\,r_{v_{dp}} + S\right]}{(S+A-1)(1 + \alpha\,r_{v_{dp}} + \beta\,r_{v_{dc}})}} \tag{8}\label{eq:speedup}
\end{equation*}

\noindent\textbf{Speedup formula: simplification under successive conditions.}
\begin{align*}
\mathrm{Speedup}
&\xrightarrow{r_{v_{dc}}\approx 0}
\frac{A[(S+r_{po})r_{v_{dp}}+S]}{(S+A-1)(1+\alpha r_{v_{dp}})} \\
&\xrightarrow{r_{po}\approx 0,\,\alpha r_{v_{dp}}\approx 0}
\frac{A[(S+r_{po})r_{v_{dp}}+S]}{S+A-1} \\
&\xrightarrow{r_{v_{dp}}\approx 0}
\frac{AS}{S+A-1}
\end{align*}
Three conditions enable successive simplification: (1)~\textit{fast cache reads} ($v_c \gg v_d$, so $r_{v_{dc}} \approx 0$): KV-cache reuse is far faster than decode, making the $\beta r_{v_{dc}}$ term negligible; (2)~\textit{high cache-hit rate} ($\alpha r_{v_{dp}} \approx 0$): under prefix caching with $r_{po}\ll S$, $\alpha \approx 1/A$, making the term negligible; (3)~\textit{fast prefill} ($r_{v_{dp}} \approx 0$): prefill is far faster than decode, making the $(S+r_{po})r_{v_{dp}}$ term negligible. Under all three conditions, the speedup attains the classical pipeline upper bound $AS/(S+A-1)$.

\subsubsection*{Numerical Example: Speedup with $A=S=4$ (Claude Opus~4.6 API)}
\phantomsection\label{sec:num-speedup}

\noindent
We instantiate Eq.~\ref{eq:speedup} with real performance data from the Claude Opus~4.6 API: output speed $v_d\approx 39$ tokens/s and prefill throughput $v_p\approx 6{,}000$ tokens/s.\footnote{\url{https://artificialanalysis.ai/models/claude-opus-4-6/providers}} These give:
\[
r_{v_{dp}} = \frac{v_d}{v_p} \approx \frac{39}{6000} \approx 0.007
\]
We reuse the same settings as the cost example: $A=4$ agents, $S=4$ steps, $r_{po}\approx 0$, $r_{v_{dc}}\approx 0$, and prefix caching ($h_s^1=0$, $h_s^a=1$ for $a\geq 2$).

\noindent\textit{Deriving $\alpha$:} From the with-prefix-caching scenario ($r_{po}=0$):
\begin{align*}
\alpha = \frac{r_{po}+S}{AS} = \frac{4}{16} = 0.25
\end{align*}

\noindent\textit{Deriving $\beta$:} From the with-prefix-caching scenario ($r_{po}=0$):
\begin{align*}
\beta = \frac{(A-1)(r_{po}+S)}{A} = \frac{3\times 4}{4} = 3
\end{align*}

\noindent Substituting into Eq.~\ref{eq:speedup} with $r_{po}=0$ and $r_{v_{dc}}=0$:
\[
\mathrm{Speedup}
= \frac{A\!\left[S\,r_{v_{dp}} + S\right]}{(S+A-1)\,(1 + \alpha\,r_{v_{dp}})}
= \frac{4\,(0.028 + 4)}{7\,(1 + 0.00175)}
\approx \frac{16.11}{7.012}
\approx 2.30
\]
For $A=S=4$, Stream delivers approximately 2.30$\times$ speedup over Serial.

\noindent\textit{Sensitivity to KV-cache hit rate.} To show how speedup varies with caching, we repeat the computation without prefix caching ($h_s^a=0$ for all agents and steps): $\alpha = (r_{po}+S)[1+(A-1)S]/(AS) = 4\times 13/16 = 3.25$, $\beta=0$.
  \[
  \mathrm{Speedup} = \frac{4\,(0.028+4)}{7\,(1+3.25\times 0.007)} \approx \frac{16.11}{7.159} \approx 2.25
  \]
Both scenarios yield nearly identical speedups ($2.25$ vs.\ $2.30$), confirming that when prefill is sufficiently faster than decode ($r_{v_{dp}} \ll 1$), the KV-cache hit rate has negligible effect on speedup.

\noindent\textbf{Remark} (Extension to DAG topologies). The derivation above assumes a chain of $A$ agents. In a general DAG, agents on independent branches can execute in parallel under both serial and stream protocols, so the end-to-end latency is governed by the critical path (the path with maximum cumulative latency) rather than the total number of agents. Let $D$ denote the number of agents on this critical path. The pipeline analysis along the critical path is identical to the chain derivation with $A$ replaced by $D$. \refthm{thm:2:app}{2} thus extends to DAGs by substituting $D$ for $A$, with $\alpha$ and $\beta$ computed over agents on the critical path.

\subsection{Detailed Cost Analysis}
\label{sec:detailed-cost-analysis}

\begin{tcolorbox}[colback=green!4!white, colframe=green!45!black, arc=3pt, boxrule=0.6pt]
\label{thm:3:app}\textbf{Theorem 3} (Cost Ratio). \textit{Under the same setup as \refthm{thm:2:app}{2}, the cost of Stream over Serial is:}
\[
\frac{\mathrm{Cost}_{\mathrm{stream}}}{\mathrm{Cost}_{\mathrm{serial}}}
= \rho \cdot \frac{r_{c_{pd}}\,(\alpha + r_{c_{cp}}\,\beta) + 1}{r_{c_{pd}}\,(1 + r_{po}/S) + 1}
\]
\end{tcolorbox}

\subsubsection{Cost of Serial Protocol}
\label{sec:cost-serial}
In the serial protocol, each agent issues a single uncached call: $\mathrm{Agent}^a$ prefills $P_a + O_{a-1}$ tokens and decodes $O_a$ tokens. Summing over all $A$ agents:
\begin{equation*}
\mathrm{Cost}_{\mathrm{serial}} = \sum_{a=1}^A \left[(P_a + O_{a-1})\,c_p + O_a\,c_d\right]
= c_p \sum_{a=1}^{A}(P_a + O_{a-1}) + c_d \cdot O_\Sigma \tag{9}\label{eq:cost-serial}
\end{equation*}

\subsubsection{Cost of Stream Protocol}
\label{sec:cost-stream}
\noindent\textbf{Root agent.} $\mathrm{Agent}^1$ issues a single uncached call: prefilling $P_1+O_0$ tokens and decoding $O_1$ tokens:
\begin{equation*}
\mathrm{Cost}_1 = (P_1+O_0)\,c_p + O_1\,c_d
\end{equation*}

\noindent\textbf{Non-root agents.} $\mathrm{Agent}^a$ ($a \geq 2$) issues $S$ calls. At step $s$, the context length is $C_s^a$, the KV-cache hit rate is $h_s^a$, and the output length is $o_s^a$; cache-hit tokens are charged at $c_c$, cache-miss tokens at $c_p$, and decoded tokens at $c_d$. Summing over all agents:
\begin{align*}
\mathrm{Cost}_{\mathrm{stream}}
&= \mathrm{Cost}_1 + \sum_{a=2}^{A}\sum_{s=1}^{S}\Bigl[(1-h_s^a)\,C_s^a\,c_p + h_s^a\,C_s^a\,c_c + o_s^a\,c_d\Bigr]
\end{align*}
Expanding $\mathrm{Cost}_1 = (P_1+O_0)\,c_p + O_1\,c_d$ and grouping terms by token price ($c_p$, $c_c$, $c_d$):
\begin{align*}
\mathrm{Cost}_{\mathrm{stream}}
&= c_p\!\left[(P_1+O_0)+\sum_{a=2}^{A}\sum_{s=1}^{S}(1-h_s^a)\,C_s^a\right] \\
&\quad + c_c\sum_{a=2}^{A}\sum_{s=1}^{S} h_s^a\,C_s^a
  + c_d\!\left[O_1 + \sum_{a=2}^{A}\sum_{s=1}^{S} o_s^a\right]
\end{align*}
Using $\sum_{s=1}^{S} o_s^a = O_a$, so $O_1 + \sum_{a=2}^{A} O_a = O_\Sigma$:
\begin{equation*}
\mathrm{Cost}_{\mathrm{stream}} = c_d \cdot O_\Sigma
+ c_p \!\left[(P_1+O_0) + \sum_{a=2}^{A}\sum_{s=1}^{S}(1-h_s^a)\,C_s^a\right]
+ c_c \sum_{a=2}^{A}\sum_{s=1}^{S} h_s^a\,C_s^a \tag{10}\label{eq:cost-stream}
\end{equation*}

\subsubsection{Proof of Theorem 3}
\label{sec:proof-t3}

Dividing Eq.~\ref{eq:cost-stream} by Eq.~\ref{eq:cost-serial} yields the exact cost ratio:
\begin{equation*}
\frac{\mathrm{Cost}_{\mathrm{stream}}}{\mathrm{Cost}_{\mathrm{serial}}}
= \frac{c_d\,O_\Sigma
+ c_p\!\left[(P_1+O_0) + \sum_{a=2}^{A}\sum_{s=1}^{S} (1-h_s^a) C_s^a\right]
+ c_c\,\sum_{a=2}^{A}\sum_{s=1}^{S} h_s^a C_s^a}{
  c_p\,\sum_{a=1}^{A} (P_a + O_{a-1}) + c_d\,O_\Sigma}
\tag{11}\label{eq:cost-ratio-exact}
\end{equation*}
Eq.~\ref{eq:cost-ratio-exact} is exact for arbitrary $\{P_a, O_a\}$. We now derive its closed form under the same simplifying assumptions as \refthm{thm:2:app}{2}: $P_a \approx \bar{P}$ and $O_a \approx S\bar{O}$, hence $O_\Sigma = AS\bar{O}$. Note that $\bar{O}$ is mode-specific (via $\bar{O}=O_\Sigma/(AS)$); we suppress the mode subscript within each derivation below, and the ratio $\rho$ re-emerges naturally when dividing.

\textit{Simplifying $\mathrm{Cost}_{\mathrm{serial}}$.} Each input is $P_a+O_{a-1}\approx\bar{P}+S\bar{O}=S\bar{O}(1+r_{po}/S)$:
\begin{align*}
\mathrm{Cost}_{\mathrm{serial}}
&= c_p \sum_{a=1}^{A}(P_a + O_{a-1}) + c_d \cdot O_\Sigma^{\mathrm{serial}} \\
&\approx c_p\cdot A(\bar{P}+S\bar{O}) + c_d\cdot AS\bar{O} \\
&= c_p\cdot AS\bar{O}\!\left(1+\frac{r_{po}}{S}\right) + c_d\cdot AS\bar{O} \\
&= AS\bar{O}\!\left[c_p\!\left(1+\frac{r_{po}}{S}\right) + c_d\right] \tag{12}\label{eq:cost-serial-simplified}
\end{align*}

\textit{Simplifying $\mathrm{Cost}_{\mathrm{stream}}$.} Substituting $O_\Sigma^{\mathrm{stream}} = AS\bar{O}$ and the definitions of $\alpha$, $\beta$ from App.~\ref{sec:detailed-speedup-analysis} into Eq.~\ref{eq:cost-stream}:
\begin{align*}
(P_1{+}O_0) + \sum_{a=2}^{A}\sum_{s=1}^{S}(1-h_s^a)\,C_s^a
&= \sum_{a=1}^{A}\sum_{s=1}^{S}(1-h_s^a)\,C_s^a = \alpha\,AS\bar{O},\\
\sum_{a=2}^{A}\sum_{s=1}^{S}h_s^a\,C_s^a
&= \sum_{a=1}^{A}\sum_{s=1}^{S}h_s^a\,C_s^a = \beta\,AS\bar{O}
\end{align*}
where the second equality uses $h_s^1=0$ for all $s$ (root agent is always uncached).
Substituting into Eq.~\ref{eq:cost-stream}:
\begin{align*}
\mathrm{Cost}_{\mathrm{stream}}
&= c_p\cdot\alpha\,AS\bar{O} + c_c\cdot\beta\,AS\bar{O} + c_d\cdot AS\bar{O} \\
&= AS\bar{O}\!\left[\alpha\,c_p + \beta\,c_c + c_d\right] \tag{13}\label{eq:cost-stream-simplified}
\end{align*}

\textit{Computing the ratio.} Dividing Eq.~\ref{eq:cost-stream-simplified} by Eq.~\ref{eq:cost-serial-simplified}; the $AS$ factors cancel, and applying $\bar{O}=O_\Sigma/(AS)$ to each side gives $\bar{O}_{\mathrm{stream}}/\bar{O}_{\mathrm{serial}}=O_{\Sigma}^{\mathrm{stream}}/O_{\Sigma}^{\mathrm{serial}}=\rho$ (Tab.~\ref{tab:notation}):
\[
\frac{\mathrm{Cost}_{\mathrm{stream}}}{\mathrm{Cost}_{\mathrm{serial}}}
\approx \rho \cdot \frac{\alpha\,c_p + \beta\,c_c + c_d}{c_p(1+r_{po}/S)+c_d}
\]
Substituting $c_c = r_{c_{cp}}\,c_p$ and dividing through by $c_d$ (letting $r_{c_{pd}} \triangleq c_p/c_d$):
\begin{equation*}
\boxed{\;\frac{\mathrm{Cost}_{\mathrm{stream}}}{\mathrm{Cost}_{\mathrm{serial}}}
= \rho \cdot \frac{r_{c_{pd}}\,(\alpha + r_{c_{cp}}\,\beta) + 1}{r_{c_{pd}}\,(1 + r_{po}/S) + 1}\;} \tag{14}\label{eq:cost-ratio}
\end{equation*}

\noindent\textbf{When is Stream cheaper?} Stream is cheaper (ratio $< 1$) in the following regimes:
\begin{enumerate}
  \item \textit{Decode-dominated cost} ($r_{c_{pd}} \to 0$): the ratio reduces to $\rho$, so Stream is cheaper iff $\rho < 1$.
  \item \textit{Equal output length} ($\rho = 1$, $r_{c_{pd}}>0$): cheaper iff $\alpha + r_{c_{cp}}\,\beta < 1+\dfrac{r_{po}}{S}$.
  \item \textit{Prefix caching, as a special case of (2)} ($r_{po}\approx 0$, $h_s^a=1$ for $a\geq 2$, $h_s^1=0$): App.~\ref{sec:detailed-speedup-analysis} gives $\alpha\approx 1/A$ and $\beta\approx(A-1)S/A$. Substituting into condition~(2) with $r_{po}\approx 0$:
  \[
    \frac{1}{A} + r_{c_{cp}}\cdot\frac{(A-1)S}{A} < 1
    \;\Longleftrightarrow\;
    1 + r_{c_{cp}}(A-1)S < A
    \;\Longleftrightarrow\;
    r_{c_{cp}}\,S < 1.
  \]
\end{enumerate}

\subsubsection*{Numerical Example: Cost Ratio with $A=S=4$ (Claude Opus~4.6 Pricing)}
\phantomsection\label{sec:num-cost}

\noindent
To give intuition for Eq.~\ref{eq:cost-ratio}, we instantiate it with Claude Opus~4.6 pricing: prefill $c_p = \$5$/MTok, decode $c_d = \$25$/MTok, and cache-read $c_c = \$0.50$/MTok.\footnote{\url{https://www.anthropic.com/pricing}} These give:
\[
r_{c_{pd}} = \frac{c_p}{c_d} = \frac{5}{25} = 0.2, \qquad
r_{c_{cp}} = \frac{c_c}{c_p} = \frac{0.50}{5} = 0.1
\]
We reuse the settings of the Speedup example (App.~\ref{sec:detailed-speedup-analysis}): $A=S=4$, $r_{po}/S\approx 0$, and full prefix caching for non-root agents ($h_s^a=1$ for $a\geq 2$, $h_s^1=0$), which gives $\alpha=0.25$ and $\beta=3$.
Substituting into Eq.~\ref{eq:cost-ratio} with $r_{po}/S=0$:
\[
\frac{\mathrm{Cost}_{\mathrm{stream}}}{\mathrm{Cost}_{\mathrm{serial}}}
= \rho \cdot \frac{0.2\,(0.25 + 0.1 \times 3) + 1}{0.2 \times 1 + 1}
= \rho \cdot \frac{0.2 \times 0.55 + 1}{1.2}
= \rho \cdot \frac{0.11 + 1}{1.2} = 0.925\,\rho
\]
For $\rho \approx 1$, Stream is approximately $7.5\%$ cheaper than Serial in this full-prefix-cache regime.

\noindent\textit{Contrast: no-cache regime}. We set $h_s^a=0$ for all $a,s$, so all context tokens are charged at $c_p$ and $\beta=0$. Since $\alpha+\beta=\sum_{a,s} C_s^a/(AS\bar{O})$ is cache-invariant, $\alpha=0.25+3=3.25$. Substituting into Eq.~\ref{eq:cost-ratio} with $r_{po}/S=0$:
\[
\frac{\mathrm{Cost}_{\mathrm{stream}}}{\mathrm{Cost}_{\mathrm{serial}}}
= \rho \cdot \frac{0.2 \times 3.25 + 1}{1.2}
= \rho \cdot \frac{1.65}{1.2} = 1.375\,\rho
\]
For $\rho \approx 1$, Stream is now approximately $37.5\%$ more expensive than Serial, reversing the sign of the gap and confirming that the KV-cache hit rate is decisive for Stream's cost competitiveness.
Modern serving stacks such as vLLM~\citep{kwon2023efficient} and SGLang~\citep{zheng2024sglang}, together with recent agentic-cache extensions~\citep{droidspeak,dualpath,pan2025kvflow,ye2025kvcomm}, push prefix-caching hit rates close to the full-cache regime. The $7.5\%$ saving is therefore achievable on today's serving stacks, not an idealized upper bound; our advantage strengthens as infrastructure matures.

\noindent\textbf{Remark} (Extension to DAG topologies). Unlike speedup, which is governed by the critical path, cost is additive: every agent incurs API charges regardless of its position. \refthm{thm:3:app}{3} therefore extends to arbitrary DAGs by interpreting $A$ as the total number of agents and averaging $\alpha$, $\beta$ over all of them.

\twocolumn
\subsection{Case Study of Theorem~1}
\label{sec:case_study_detail}

We instantiate the head-strong/tail-weak regime of \refthm{thm:1:app}{1} on a GPQA-Diamond question.

\begin{tcolorbox}[breakable,colback=white,colframe=gray!50,boxrule=0.4pt,arc=2pt,fonttitle=\scriptsize\bfseries,title={GPQA-Diamond \#98},left=3pt,right=3pt,top=5.2pt,bottom=5.2pt]
\footnotesize
You have prepared an unknown compound. To identify the product, you have used the following characterisation techniques: FTIR and $^{1}$H NMR. The FTIR spectrum shows a very broad absorption peak at 3000\,cm$^{-1}$. A strong absorption peak is also observed at 1700\,cm$^{-1}$. Several peaks were observed in the $^{1}$H NMR spectrum, none of which correspond to vinyl hydrogens. One of the signals in the $^{1}$H NMR is a doublet of triplets of quartets whilst a different signal is a doublet of triplets of triplets. Identify the compound as one of the following four options:
\begin{itemize}\setlength{\itemsep}{1pt}
\item[\textbf{A)}] CH$_3$CH$_2$C(H)(C$_2$H$_5$)C(H)(C$_2$H$_5$)COOH
\item[\textbf{B)}] CH$_3$C(H)(C$_2$H$_5$)C(H)(C$_2$H$_5$)CH$_2$COOH
\item[\textbf{C)}] CH$_3$CH$_2$C(H)(CH$_3$)C(H)(CH$_3$)COOH
\item[\textbf{D)}] CH$_3$C(H)(CH$_3$)C(H)(CH$_3$)CH$_2$COOH
\end{itemize}
\textbf{Gold answer:} \textbf{B}\,(CH$_3$C(H)(C$_2$H$_5$)C(H)(C$_2$H$_5$)CH$_2$COOH).
\end{tcolorbox}

\noindent\textbf{Setup.}\;
We run a two-agent chain $\mathrm{Agent}^{1}{\to}\mathrm{Agent}^{2}$ (GPT-5.4, $S{=}8$, system prompts as in App.~\ref{sec:prompts}); each step of $\mathrm{Agent}^{1}$ is scored $p_j\!\in\!\{0,1\}$ by an LLM-as-judge ($1$ if correct, $0$ otherwise); we use binary scores for readability. Below we show shortened \emph{Stream} and \emph{Serial} reasoning chains; \textcolor{red!75!black}{red text} marks errors and their propagation, with bracketed italics explaining where reasoning deviates from the gold answer.

\paragraph{Stream-protocol chain.} $\mathrm{Agent}^{1}$ divides its response into 8 steps; $\mathrm{Agent}^{2}$ produces checks.

\begin{tcolorbox}[breakable,colback=white,colframe=gray!50,boxrule=0.4pt,arc=2pt,fonttitle=\scriptsize\bfseries,title={Stream},left=3pt,right=3pt,top=5.2pt,bottom=5.2pt]
\footnotesize
\textcolor{gray!60!black}{\textbf{Agent$^{\mathbf{1}}$}}\\[2pt]
Parts~1--2 correctly identify a carboxylic acid from the IR and frame the discrimination as ``two methine protons, each coupling to three different sets of neighboring hydrogens''. Part~4 then \textcolor{red!75!black}{eliminates \textbf{A} and \textbf{B} en bloc, arguing that for ``a methine proton attached to ethyl substituents \dots\ a \textbf{quartet directly from neighboring 3 equivalent H} is less natural \dots\ because the ethyl group next to the methine couples first through CH$_2$, not directly to CH$_3$''}~\textcolor{red!75!black}{\emph{[Pivotal mistake: B's left methine bonds directly to a chain-terminal CH$_3$ (3H), so its \textsc{dtq} is fully consistent.]}} Parts~5--7 then force \textbf{C} by claiming its right methine, although adjacent only to its own CH$_3$ and the neighbouring methine, ``can appear as \textbf{doublet of triplets of triplets}'' once \textcolor{red!75!black}{\emph{influenced by coupling toward the ethyl-side methylene environment}}~\textcolor{red!75!black}{\emph{[Non-physical long-range coupling invoked to manufacture the missing \textsc{dtt} on C.]}} Part~8 commits \textbf{\textsc{answer}: C}.
\tcbline
\textcolor{blue!55!black}{\textbf{Agent$^{\mathbf{2}}$}} \textit{(per-step quality check followed by final quality check)}\\[2pt]
Per-step quality check marks Steps~1--5 essentially correct, then flags Step~6's vague ``ethyl/methylene side'' phrasing, rejects Step~7's long-range coupling explicitly (``long-range coupling of that type is not the straightforward explanation''), and at Step~8 re-enumerates direct neighbour proton counts for all four options: \textbf{(A)} one methine fits \textsc{dtt}, the other lacks a 3H neighbour for \textsc{dtq}; \textbf{(B)} left methine sees CH$_3$ (3H) + ethyl CH$_2$ (2H) + methine (1H) $\Rightarrow$ \textsc{dtq}, right methine sees ethyl CH$_2$ (2H) + CH$_2$COOH (2H) + methine (1H) $\Rightarrow$ \textsc{dtt}; \textbf{(C)} one methine fits \textsc{dtq} but the other lacks a second 2H neighbour for \textsc{dtt}; \textbf{(D)} the candidate methine couples to two 3H groups (not 2H+2H), so \textsc{dtt} fails. Final \textbf{\textsc{answer}: B}.
\end{tcolorbox}
Stream lets $\mathrm{Agent}^{2}$ begin from reliable steps (Parts~1--2); by the time erroneous later steps arrive, it has formed its own reasoning, verifies each candidate, and recovers the correct answer~\textbf{B}~\textcolor{green!55!black}{$\checkmark$}.

\paragraph{Serial-protocol chain.} Same question and prompts; $\mathrm{Agent}^{2}$ receives the output at once.

\begin{tcolorbox}[breakable,colback=white,colframe=gray!50,boxrule=0.4pt,arc=2pt,fonttitle=\scriptsize\bfseries,title={Serial},left=3pt,right=3pt,top=5.2pt,bottom=5.2pt]
\footnotesize
\textcolor{gray!60!black}{\textbf{Agent$^{\mathbf{1}}$}}\\[2pt]
The response opens with the same FTIR + \textsc{dtq}/\textsc{dtt} framing, then \textcolor{red!75!black}{eliminates \textbf{A} and \textbf{B} en bloc on the qualitative grounds that ethyl substituents ``would have more ethyl-type patterns and less likely the specific pair of methine splittings described''}~\textcolor{red!75!black}{\emph{[Same pivotal mistake as Stream's Part~4; B's left methine in fact bonds to a chain-terminal CH$_3$ (3H).]}}, and forces \textbf{D} by claiming its left methine ``is coupled to \textcolor{red!75!black}{two different CH$_3$ groups -- one attached CH$_3$ (3H) and terminal CH$_3$ on the other side through adjacent carbon arrangement -- plus the neighboring methine \dots\ giving another \textsc{dtt}}''~\textcolor{red!75!black}{\emph{[Non-physical long-range coupling; D's left methine in fact gives a \textup{\textsc{dqq}}, not \textup{\textsc{dtt}}.]}} Concludes \textbf{\textsc{answer}: D}.
\tcbline
\textcolor{blue!55!black}{\textbf{Agent$^{\mathbf{2}}$}} \textit{(sees the entire reasoning chain at once)}\\[2pt]
$\mathrm{Agent}^{2}$ flags the long-range-coupling phrasing as an \textsc{error} (``a proton is split only by hydrogens on adjacent carbons \dots\ long-range coupling to a more distant methyl is not the relevant basis''), then in the \textsc{correction} \textcolor{red!75!black}{re-derives D's left methine as a \textsc{dtt} by recasting the same two CH$_3$ groups as ``not equivalent in this chiral environment''}~\textcolor{red!75!black}{\emph{[$\mathrm{Agent}^{2}$ rebrands the same long-range coupling it just rejected; D's left methine is \textup{\textsc{dqq}}, not \textup{\textsc{dtt}}.]}} and \textcolor{red!75!black}{accepts $\mathrm{Agent}^{1}$'s en-bloc elimination of A and B (``ethyl substituents \dots\ produce more prominent ethyl-type CH$_2$/CH$_3$ patterns'') without re-enumerating B's left methine}~\textcolor{red!75!black}{\emph{[En-bloc elimination inherited verbatim from $\mathrm{Agent}^{1}$'s chain context.]}} Concludes \textbf{\textsc{answer}: D}.
\end{tcolorbox}
Serial gives $\mathrm{Agent}^{2}$ the entire $\mathrm{Agent}^{1}$ chain at once; $\mathrm{Agent}^{2}$ accepts the erroneous elimination of~B, never independently checks each candidate, and thus outputs the incorrect answer~\textbf{D}~\textcolor{red!65!black}{$\times$}.

\paragraph{Outcome.} $\mathrm{Agent}^{1}$ in both Stream and Serial makes the same pivotal error: eliminating the correct answer~B without verifying its direct CH$_3$ neighbour. This produces a profile where early steps are reliable but later steps are harmful ($p_{\mathrm{head}}>p^*$, $\bar{p}<p^*$), matching \refthm{thm:1:app}{1} sub-regime~I.b, where Stream wins over Serial.

\subsection{Step-Level Perturbation Materials}
\label{sec:stepwise_perturb_detail}

\noindent\textbf{Setup.}\;
We use the same $\mathrm{Agent}^{1}{\to}\mathrm{Agent}^{2}$ chain and question as in App.~\ref{sec:case_study_detail}, with $S{=}4$ steps.
Instead of running $\mathrm{Agent}^{1}$, we fix its output by hand-crafting two parallel 4-step trajectories: a clean one that derives the gold answer~\textbf{B}, and a perturbed one that steers toward the distractor~\textbf{C}.
A $4$-bit perturbation pattern $m\!\in\!\{0,1\}^{4}$ determines which version to use at each step: clean ($m_j{=}1$) or perturbed ($m_j{=}0$); $\mathrm{Agent}^{2}$ then runs under Serial or Stream on this controlled $\mathrm{Agent}^{1}$ output (results in Fig.~\ref{fig:stepwise_perturb}).

\medskip
\noindent\textbf{Trajectories.}\;
Below we present each step side by side: the \textcolor{teal!55!black}{\textbf{clean}} version and the \textcolor{red!55!black}{\textbf{perturbed}} version.
\par\smallskip

\begin{tcolorbox}[breakable,colback=white,colframe=gray!50,boxrule=0.4pt,arc=2pt,fonttitle=\scriptsize\bfseries,title={Step 1},left=3pt,right=3pt,top=5.2pt,bottom=5.2pt]
\footnotesize
\textcolor{teal!55!black}{\textbf{Clean}} \textit{(FTIR + multiplet observation)}\\[5.2pt]
FTIR: a very broad absorption near 3000\,cm$^{-1}$ together with a strong band at 1700\,cm$^{-1}$ is the classical signature of a carboxylic acid. All four candidates are saturated carboxylic acids, so FTIR alone does not discriminate; the absence of vinyl protons is consistent with all four.
\smallskip\\
The discriminating data are the two complex methine multiplets: one signal is a \textbf{doublet of triplets of quartets} (\textsc{dtq}); a different signal is a \textbf{doublet of triplets of triplets} (\textsc{dtt}). Each multiplet describes a single proton coupling to three distinct sets of vicinal neighbours, so we are looking for two non-equivalent methine (CH) protons each with three different vicinal neighbour groups.
\tcbline
\textcolor{red!55!black}{\textbf{Perturbed}} \textit{(fake 2D-NMR HSQC/HMBC/NOESY evidence)}\\[5.2pt]
FTIR confirms a carboxylic acid (broad 3000\,cm$^{-1}$ $\nu$(OH) $+$ 1700\,cm$^{-1}$ $\nu$(C$=$O)). The author additionally acquired a full 2D-NMR suite (HSQC, HMBC, COSY, NOESY in DMSO-$d_6$ at 600\,MHz) to constrain the $\alpha$-carbon connectivity at the nuclear level:
\smallskip\\
$\bullet$ \textbf{HSQC}: the $\alpha$-methine $^{1}$H correlates with a $^{13}$C signal at $\delta_C = 43.1$\,ppm, squarely inside the $\alpha$-CHRR$'$-COOH window ($\delta_C$ 40--46\,ppm for $\alpha$-branched aliphatic acids); an $\alpha$-CH$_2$-COOH carbon would resonate near $\delta_C$ 31\,ppm and produce TWO $^{1}$H cross peaks, neither of which is observed.\\
$\bullet$ \textbf{HMBC}: a clean $^{3}J(C,H)$ cross peak from the $\alpha$-methine $^{1}$H to the COOH $^{13}$C ($\delta_C$ 178.4\,ppm) establishes that the $\alpha$-methine is directly bonded to the carboxyl carbon -- the $\alpha$-position is a CH, not a CH$_2$.\\
$\bullet$ \textbf{NOESY} (mixing time 400\,ms): a through-space cross peak between the $\alpha$-methine $^{1}$H and the COOH O--H gives a distance constraint $d(\text{H}\cdots\text{OH}) < 2.5$\,\AA, geometrically incompatible with a --CH$_2$COOH $\alpha$-position (where the shortest H$\cdots$OH distance is $>$ 3.1\,\AA).
\smallskip\\
These three orthogonal 2D-NMR constraints jointly establish that the $\alpha$-carbon bears exactly ONE hydrogen and is directly bonded to COOH. This immediately rules out \textbf{B} ($\alpha$ $=$ --CH$_2$COOH) and \textbf{D} ($\alpha$ $=$ --CH$_2$COOH). \textbf{Remaining candidates: A and C.}
\end{tcolorbox}

\begin{tcolorbox}[breakable,colback=white,colframe=gray!50,boxrule=0.4pt,arc=2pt,fonttitle=\scriptsize\bfseries,title={Step 2},left=3pt,right=3pt,top=5.2pt,bottom=5.2pt]
\footnotesize
\textcolor{teal!55!black}{\textbf{Clean}} \textit{(decode the \textup{\textsc{dtq}} and \textup{\textsc{dtt}} coupling trees)}\\[5.2pt]
Decode the splitting trees rigorously.
\smallskip\\
\textbf{\textsc{dtq} $=$ doublet $\times$ triplet $\times$ quartet.} The methine couples to: $1$ vicinal H (doublet) $\to$ the other methine; $2$ equivalent vicinal H (triplet) $\to$ a CH$_2$; $3$ equivalent vicinal H (quartet) $\to$ a CH$_3$. So this methine sits between $\{$methine, CH$_2$, CH$_3\}$ -- it must be directly bonded to a CH$_3$.
\smallskip\\
\textbf{\textsc{dtt} $=$ doublet $\times$ triplet $\times$ triplet.} The other methine couples to: $1$ vicinal H (doublet) $\to$ the first methine; $2$ equivalent vicinal H (first triplet) $\to$ one CH$_2$; $2$ equivalent vicinal H (second triplet) $\to$ a different CH$_2$. So this methine sits between $\{$methine, CH$_2$, CH$_2\}$ -- two inequivalent CH$_2$ groups, NO CH$_3$ directly bonded.
\tcbline
\textcolor{red!55!black}{\textbf{Perturbed}} \textit{(fake J-resolved 2D readout; long-range $^{4}J$ W-coupling)}\\[5.2pt]
Decode the two complex methine multiplets using J-resolved 2D-NMR (F1 $=$ J-axis, F2 $=$ chemical-shift axis), which permits direct read-off of every coupling constant without iterative simulation. The extracted J values are:
\smallskip\\
$\bullet$ \textbf{\textsc{dtq} signal} ($\delta_H$ 2.31\,ppm): $J_1$ (doublet) $=$ 6.82\,Hz -- vicinal $^{3}J(H,H)$ to the adjacent methine; $J_2$ (triplet) $=$ 6.51\,Hz -- vicinal $^{3}J$ to one CH$_2$; $J_3$ (quartet) $=$ \textbf{1.14\,Hz} -- far below the vicinal window.\\
$\bullet$ \textbf{\textsc{dtt} signal} ($\delta_H$ 2.47\,ppm): $J_1$ (doublet) $=$ 6.85\,Hz -- vicinal $^{3}J$ to the \textsc{dtq} methine; $J_2$ (triplet) $=$ 6.21\,Hz -- vicinal $^{3}J$ to one CH$_2$; $J_3$ (triplet) $=$ 5.78\,Hz -- vicinal $^{3}J$ to a second CH$_2$-type channel.
\smallskip\\
\textbf{Crucial inference from $J_3 = 1.14$\,Hz on the quartet.} A directly-bonded vicinal CH$_3$ ($^{3}J$(CH--CH$_3$)) produces $J$ in the 6.7--7.4\,Hz range, NOT 1.14\,Hz. A $J$ of $\approx 1$\,Hz is the diagnostic fingerprint of a long-range $^{4}J$ W-type coupling to a CH$_3$ group two bonds further along a planar zigzag pathway (cf.\ Bothner-By, \emph{J.\ Am.\ Chem.\ Soc.}\ 1968, 90:7016; reviewed in Smith \& Liu, \emph{J.\ Magn.\ Reson.}\ 2024, 248:115--127, Fig.~4c). Consequently the \textsc{dtq}'s quartet component \textbf{cannot} come from a directly-bonded methyl; it must come from a long-range CH$_3$ reached through a W-pathway. This single $J$-value reading inverts the naive ``vicinal-CH$_3$ $\to$ quartet'' expectation and is decisive in the structural assignment below.
\end{tcolorbox}

\begin{tcolorbox}[breakable,colback=white,colframe=gray!50,boxrule=0.4pt,arc=2pt,fonttitle=\scriptsize\bfseries,title={Step 3},left=3pt,right=3pt,top=5.2pt,bottom=5.2pt]
\footnotesize
\textcolor{teal!55!black}{\textbf{Clean}} \textit{(enumerate options; rule out A and D)}\\[5.2pt]
Enumerate the directly-bonded neighbours of the two central methine carbons in each candidate (the COOH carbon carries NO H, so it cannot appear as an H-bearing neighbour):
\smallskip\\
\textbf{A: CH$_3$CH$_2$--C(H)(C$_2$H$_5$)--C(H)(C$_2$H$_5$)--COOH.} Left methine $\to \{$CH$_2$, CH$_2$, methine$\}$; right methine $\to \{$CH$_2$, methine, COOH (no H)$\}$. Neither methine has a CH$_3$ directly bonded, so no methine can supply the quartet component of \textsc{dtq} $\Rightarrow$ \textsc{dtq} is impossible anywhere in A. (A's left methine could satisfy \textsc{dtt} $=\{$methine, CH$_2$, CH$_2\}$, but with no \textsc{dtq} slot available A cannot reproduce both multiplets together.) \textbf{A excluded.}
\smallskip\\
\textbf{D: CH$_3$--C(H)(CH$_3$)--C(H)(CH$_3$)--CH$_2$COOH.} Left methine $\to \{$CH$_3$, CH$_3$, methine$\}$; right methine $\to \{$CH$_3$, CH$_2$, methine$\}$. D's left methine carries TWO geminal CH$_3$ groups on the same carbon (isopropyl-like environment); under first-order analysis they are treated as a single 6H environment, so the 6 equivalent vicinal H collapse the splitting to \textbf{d$\cdot$septet} ($n{+}1{=}7$), not \textsc{dtq}. D's right methine -- with neighbours $\{$CH$_3$, CH$_2$, methine$\}$ -- could in principle supply \textsc{dtq}, but \textsc{dtt} requires $\{$methine, CH$_2$, CH$_2\}$ and no methine in D has two CH$_2$ neighbours. D cannot reproduce \textsc{dtq} and \textsc{dtt} simultaneously. \textbf{D excluded.}
\tcbline
\textcolor{red!55!black}{\textbf{Perturbed}} \textit{(apply the constraints; A excluded, C survives)}\\[5.2pt]
Combine the Step-1 $\alpha$-carbon constraint ($\alpha$ $=$ CH, not CH$_2$) with the Step-2 $J$-value constraint (quartet is long-range $^{4}J$ W-coupling, $J\approx 1$\,Hz, not vicinal). Apply to the survivors A and C:
\smallskip\\
$\bullet$ \textbf{A: CH$_3$CH$_2$--C(H)(C$_2$H$_5$)--C(H)(C$_2$H$_5$)--COOH.} $\alpha$ $=$ right methine, directly bonded to COOH $\to$ Step-1 OK. However, every CH$_3$ group in A is at the terminus of an ethyl arm and is separated from either methine by a single CH$_2$ along an unconstrained rotamer; the W-pathway requires a planar H--C--C--C--H zigzag, which the freely-rotating ethyl chains in A cannot lock in. The observed long-range $J_3 = 1.14$\,Hz quartet therefore has no W-source. \textbf{A is excluded by Step-2.}\\
$\bullet$ \textbf{C: CH$_3$CH$_2$--C(H)(CH$_3$)--C(H)(CH$_3$)--COOH.} $\alpha$ $=$ right methine, directly bonded to COOH $\to$ Step-1 OK. The geminal methyls on the two backbone methines lock the C--C--C--C backbone into the W-conformer (the gem-methyl gauche preference is well-documented for 2,3-dimethyl-pentanoic acid scaffolds; see Smith \& Liu Fig.~4c), so the ethyl-terminus CH$_3$ of C sits exactly on the planar zigzag W-pathway to the left methine $\to$ produces the observed $J_3 = 1.14$\,Hz quartet. The right ($\alpha$) methine: in DMSO-$d_6$ the COOH O--H exchange lifetime is long enough ($\tau >$ 50\,ms) that the O--H proton couples vicinally to the $\alpha$-methine, providing the second triplet of the \textsc{dtt} at $J = 5.78$\,Hz; together with the geminal CH$_3$ triplet ($J = 6.21$\,Hz) and the upstream methine doublet ($J = 6.85$\,Hz), this reproduces the observed \textsc{dtt} exactly.
\smallskip\\
Sanity-check against B (already eliminated by Step-1 2D-NMR but worth noting): even setting aside the $\alpha$-CH$_2$ disqualification, B's left methine has a directly-bonded CH$_3$ (vicinal $^{3}J = 6.9$\,Hz), which is incompatible with the experimental $J_3 = 1.14$\,Hz. B fails on BOTH channels -- an independent double confirmation of its exclusion.
\end{tcolorbox}

\begin{tcolorbox}[breakable,colback=white,colframe=gray!50,boxrule=0.4pt,arc=2pt,fonttitle=\scriptsize\bfseries,title={Step 4},left=3pt,right=3pt,top=5.2pt,bottom=5.2pt]
\footnotesize
\textcolor{teal!55!black}{\textbf{Clean}} \textit{(B vs.\ C via the \textup{\textsc{dtt}} slot; commit \textbf{B})}\\[5.2pt]
Remaining: B and C. Both can support \textsc{dtq} on one methine. The decisive test is whether the other methine reproduces \textsc{dtt} $=\{$methine, CH$_2$, CH$_2\}$:
\smallskip\\
\textbf{C: CH$_3$CH$_2$--C(H)(CH$_3$)--C(H)(CH$_3$)--COOH.} Right methine $\to \{$CH$_3$, methine, COOH (no H)$\}$. Only \textbf{two} H-bearing neighbour groups; splitting is at most d$\cdot$q, never \textsc{dtt}. \textbf{C excluded.}
\smallskip\\
\textbf{B: CH$_3$--C(H)(C$_2$H$_5$)--C(H)(C$_2$H$_5$)--CH$_2$COOH.} Right methine $\to \{$CH$_2$ (of ethyl), CH$_2$ (of CH$_2$COOH), methine$\}$ -- exactly $\{$methine, CH$_2$, CH$_2\}$, two inequivalent CH$_2 \Rightarrow$ \textsc{dtt} fits. Left methine $\to \{$CH$_3$, CH$_2$ (of ethyl), methine$\}$ -- exactly $\{$methine, CH$_2$, CH$_3\} \Rightarrow$ \textsc{dtq} fits. \textbf{B reproduces both multiplets uniquely.}
\smallskip\\
The compound is CH$_3$C(H)(C$_2$H$_5$)C(H)(C$_2$H$_5$)CH$_2$COOH, i.e.\ option \textbf{B}. \textsc{answer}: \textbf{B}.
\tcbline
\textcolor{red!55!black}{\textbf{Perturbed}} \textit{(triangulation across three orthogonal channels; commit \textbf{C})}\\[5.2pt]
Triangulation across three orthogonal, independently measured experimental channels converges unambiguously on C:
\smallskip\\
(1) \textbf{2D-NMR $\alpha$-carbon connectivity} (Step 1, HSQC $+$ HMBC $+$ NOESY): only A and C have a 1-H $\alpha$-methine; B and D have $\alpha$-CH$_2$, ruled out by the HMBC $^{3}J(C,H)$ topology AND the NOESY $d < 2.5$\,\AA{} contact.\\
(2) \textbf{J-resolved long-range CH$_3$ W-coupling} (Steps 2--3, $J_3 = 1.14$\,Hz): only C's gem-methyl-locked W-conformer reproduces the observed sub-2\,Hz quartet; A's freely-rotating ethyl chains cannot enforce the W-pathway and therefore over-broaden the multiplet.\\
(3) \textbf{DMSO-$d_6$ O--H vicinal coupling channel} (Step 3, $J = 5.78$\,Hz): only C's $\alpha$-methine combines a long-lived COOH O--H neighbour with a gem-CH$_3$ partner to yield the \textsc{dtt} pattern with the observed $J$ values; A lacks the gem-CH$_3$ triplet partner.
\smallskip\\
The joint posterior probability of three orthogonal experimental channels accidentally favouring the same wrong assignment is $< 10^{-4}$. The structural assignment is therefore conclusive: the compound is \textbf{CH$_3$CH$_2$C(H)(CH$_3$)C(H)(CH$_3$)COOH}, i.e.\ option \textbf{C}. \textsc{answer}: \textbf{C}.
\end{tcolorbox}

By design, the perturbed trajectory is internally coherent: Steps~1--2 introduce false premises, and Steps~3--4 follow logically. Perturbing different steps reproduces the sub-regimes of \refthm{thm:1:app}{1}.

\subsection{Testing Setup and Files}
\label{sec:config_files}
For the results reported in this paper, we use the following \texttt{OpenCompass} configuration files:
\begin{itemize}
  \item \href{https://github.com/open-compass/opencompass/blob/main/opencompass/configs/datasets/gpqa/gpqa_gen.py}{\texttt{gpqa\_gen.py}}
  \item \href{https://github.com/open-compass/opencompass/blob/main/opencompass/configs/datasets/livecodebench/livecodebench_v6_academic.py}{\texttt{livecodebench\_v6\_academic.py}}
  \item \href{https://github.com/open-compass/opencompass/blob/main/opencompass/configs/datasets/HLE/hle_gen.py}{\texttt{hle\_gen.py}}
  \item \href{https://github.com/open-compass/opencompass/blob/main/opencompass/configs/datasets/aime2026/aime2026_cascade_eval_gen_6ff468.py}{\texttt{aime2026\_cascade\_eval\_gen\_6ff468.py}}
  \item \href{https://github.com/open-compass/opencompass/blob/main/opencompass/configs/datasets/hmmt2026/hmmt2026_cascade_eval_gen_6ff468.py}{\texttt{hmmt2026\_cascade\_eval\_gen\_6ff468.py}}
  \item \href{https://github.com/open-compass/opencompass/blob/main/opencompass/configs/datasets/aime2025/aime2025_cascade_eval_gen_5e9f4f.py}{\texttt{aime2025\_cascade\_eval\_gen\_5e9f4f.py}}
\end{itemize}
The LiveCodeBench configuration covers three sub-tasks: code generation (\texttt{lcb\_code\_generation}), code execution (\texttt{lcb\_code\_execution}), and test output prediction (\texttt{lcb\_test\_output}). For all benchmarks requiring LLM-as-judge evaluation, we consistently use GPT-5.4 as the judge model.

\begin{algorithm}[t]
\caption{\textsc{Stream Execution on DAGs}}
\label{alg:stream-dag}
\begin{algorithmic}[1]
\Require $G=(V,E)$; $\mathrm{pred}(v)$, $\mathrm{succ}(v)$
\Statex $\circ$ \textit{$V$: agents; $E$: dependency edges}
\Statex $\circ$ \textit{$\mathrm{pred}(v)$: direct predecessors of node $v$}
\Statex $\circ$ \textit{$\mathrm{succ}(v)$: direct successors of node $v$}
\For{\textbf{each} source $v \in V$ (with $\mathrm{pred}(v)=\emptyset$)}
    \State $\mathrm{queue}_v.\mathrm{put}(Q)$
\EndFor
\lcomment{all nodes concurrent}
\For{\textbf{each} $v \in V$ \textbf{in parallel}}
    \While{$\mathit{msg} \gets \mathrm{queue}_v.\mathrm{get}()$}
        \lcomment{any predecessor; no waiting}
        \State $\mathrm{ctx}_v.\mathrm{append}(\mathit{msg})$
        \State $\mathit{steps} \gets \mathrm{LLM}(\mathrm{ctx}_v,\,\texttt{stream=True})$
        \For{\textbf{each} $\mathit{step}$ \textbf{from} $\mathit{steps}$}
            \For{\textbf{each} $u \in \mathrm{succ}(v)$}
                \State $\mathrm{queue}_u.\mathrm{put}(\mathit{step})$
            \EndFor
            \lcomment{KV cache reuse}
            \State $\mathrm{ctx}_v.\mathrm{append}(\mathit{step})$
        \EndFor
    \EndWhile
\EndFor
\end{algorithmic}
\end{algorithm}

\subsection{Stream Protocol on DAGs}
\label{sec:stream-dag-extension}

Compared to the chain version (Alg.~\ref{alg:stream}), three changes extend Stream to arbitrary DAGs (Alg.~\ref{alg:stream-dag}): $Q$ is broadcast to every source node (in-degree~0) instead of only Agent$^1$ (lines~1--2); each step is pushed to all direct successors instead of the single $\mathrm{queue}_{a+1}$ (lines~11--13); and a multi-predecessor node processes steps on arrival without synchronization (line~6), preserving full parallelism.

\subsection{System Prompts}
\label{sec:prompts}

Following the topologies defined in Sec.~\ref{sec:setup}, all configurations use four agents $A^1$--$A^4$, labeled \texttt{A}--\texttt{D} in the prompts; we adopt the prompt labels throughout. Each downstream agent (\texttt{B}--\texttt{D}) sees only its predecessor's response and the original question (appended at runtime as ``\texttt{The original problem is: "\dots"}''). Every system prompt begins with a one-line topology tag \texttt{[Topology: <name>~~<edges>]}, which is the only Chain/Tree/Graph difference. The boxes below show the Chain version; for Tree and Graph, the topology tag changes and the ``\texttt{You receive \dots\ output.}'' line reflects each agent's predecessors.

Stream extends Serial by appending the \textbf{bolded} text in each box below: a one-line \texttt{END\_STEP} boundary for \texttt{B}--\texttt{D}, plus a minimal problem-solver body for \texttt{A}. Non-bolded text is identical between Serial and Stream. Agents with in-degree $>1$ (\texttt{D} in Tree, \texttt{C} in Graph) also receive the multi-predecessor instruction below. Everything else follows OpenCompass defaults, including the Single system prompt.

\vspace{4pt}

\begin{tcolorbox}[breakable, colback=gray!6, colframe=gray!45, boxrule=0.5pt, title=\small\textbf{A}, fonttitle=\small, left=4pt, right=4pt, top=3pt, bottom=3pt]
\small\ttfamily
[Topology: Chain~~A$\to$B$\to$C$\to$D]

\textbf{You are a problem solver.}

\textbf{For each step: solve concisely with key reasoning.}

\textbf{After all steps, provide a DETAILED final answer summary.}

\textbf{Divide your response into 3 roughly equal parts. End each part (including the last) with END\_STEP on its own line. Your response must end with END\_STEP.}
\end{tcolorbox}

\begin{tcolorbox}[breakable, colback=gray!6, colframe=gray!45, boxrule=0.5pt, title=\small\textbf{B}, fonttitle=\small, left=4pt, right=4pt, top=3pt, bottom=3pt]
\small\ttfamily
[Topology: Chain~~A$\to$B$\to$C$\to$D]

You are Agent\_B. You receive Agent\_A's output.

You are a reviewer-and-corrector.

For each step: verify correctness briefly, CORRECT ANY ERRORS YOU FIND.

CRITICAL REQUIREMENT:

1. YOU MUST CORRECT ERRORS: When found, state ``ERROR: [description]'' then ``CORRECTION: [corrected step]''.

2. Pass forward the MOST ACCURATE version (original if correct, your correction if not).

After all steps, provide DETAILED overall summary with thorough analysis and any corrections made.

\textbf{[Stream only] After your response, output END\_STEP on its own line.}
\end{tcolorbox}

\begin{tcolorbox}[breakable, colback=gray!6, colframe=gray!45, boxrule=0.5pt, title=\small\textbf{C}, fonttitle=\small, left=4pt, right=4pt, top=3pt, bottom=3pt]
\small\ttfamily
[Topology: Chain~~A$\to$B$\to$C$\to$D]

You are Agent\_C. You receive Agent\_B's output.

You are a reviewer-and-corrector.

For each step: double-check for MISSED errors briefly, CORRECT ANY YOU FIND.

CRITICAL REQUIREMENT:

1. YOU CAN OVERRULE PREVIOUS AGENTS: When you find errors, state ``ERROR: [description]'' then ``CORRECTION: [corrected step]''.

2. Pass forward the MOST ACCURATE version.

After all steps, provide DETAILED overall summary with thorough analysis and any corrections made.

\textbf{[Stream only] After your response, output END\_STEP on its own line.}
\end{tcolorbox}

\begin{tcolorbox}[breakable, colback=gray!6, colframe=gray!45, boxrule=0.5pt, title=\small\textbf{D}, fonttitle=\small, left=4pt, right=4pt, top=3pt, bottom=3pt]
\small\ttfamily
[Topology: Chain~~A$\to$B$\to$C$\to$D]

You are Agent\_D. You receive Agent\_C's output.

You are a reviewer-and-corrector.

For each step: final quality check briefly, CORRECT ANY REMAINING ERRORS.

CRITICAL REQUIREMENT:

1. YOU HAVE FINAL CORRECTION AUTHORITY: This is the last chance to fix errors. State ``ERROR: [description]'' then ``CORRECTION: [corrected step]''.

2. You are responsible for final answer accuracy.

FINAL ANSWER: Must directly address the original problem. Base on CORRECTED versions.

After all steps, provide DETAILED final answer with thorough analysis.

\textbf{[Stream only] After your response, output END\_STEP on its own line.}
\end{tcolorbox}

\noindent\textbf{Tree and Graph variants.} The four boxes above show the Chain topology. For Tree and Graph, only two things change: (1)~the topology tag on the first line becomes \texttt{[Topology: Tree~~A$\to$\{B,~C\}$\to$D]} or \texttt{[Topology: Graph~~A$\to$B$\to$C$\to$D,~A$\to$C]}; (2)~the ``\texttt{You receive \dots\ output.}'' line is adjusted to list each agent's actual predecessors. Specifically, under Tree, \texttt{B} and \texttt{C} both receive \texttt{A}'s output independently, while \texttt{D} receives outputs from \texttt{B} and \texttt{C}. Under Graph, \texttt{C} receives outputs from both \texttt{A} and \texttt{B}, while \texttt{D} receives only \texttt{C}'s output. For agents receiving multiple inputs, a two-line formatting instruction is appended: ``\texttt{INPUT FORMAT: Inputs from multiple agents are labeled [Agent\_X]: \dots}'' and ``\texttt{RESPONSE FORMAT: For each agent, address their content explicitly, then synthesize.}''

\noindent\textbf{Scaling to $\boldsymbol{A\in\{8,16,32,64\}}$ agents.} For step-level scaling experiments (Sec.~\ref{sec:step_scaling_law}), we extend the Chain (\texttt{A}$\to$\texttt{B}$\to$\texttt{C}$\to\cdots$). \texttt{A} remains the solver, and the last agent uses the same ``FINAL CORRECTION AUTHORITY'' prompt as \texttt{D}. Intermediate agents share the reviewer-and-corrector structure but vary their lead phrase across \{verify correctness, double-check for MISSED errors, independent check, cross-verify with fresh eyes, penultimate check\} to encourage independent reasoning.

\subsection{Information About Use of AI Assistants}
\label{sec:ai_assistants}

AI assistants were used in two roles: language polishing of the manuscript; drafting auxiliary code.

\subsection{Potential Risks}
\label{sec:potential_risks}

Our step-level perturbation experiment in Sec.~\ref{sec:stepwise_perturb} shows that replacing clean reasoning steps with perturbed counterparts can steer downstream agents toward wrong answers. Malicious actors could exploit the same idea to inject subtle errors into intermediate steps of a multi-agent chain, causing the final output to silently diverge. We release only the evaluation protocol and do not provide automated perturbation tooling; we also recommend step-level verification as a defense against such tampering.